\documentclass[12pt]{article}
\usepackage{amsfonts,amssymb}
\usepackage{hyperref}
\usepackage{titlesec}
\usepackage{titletoc}
\usepackage{amsmath}
\numberwithin{equation}{section}
\usepackage{listings}
\usepackage{verbatim}
\usepackage{graphicx}
\usepackage{geometry}
\usepackage{mathrsfs}
\usepackage{float}
\usepackage[mathlines]{lineno}
\usepackage{amsthm}
\usepackage{bbm}
\usepackage{bm}

\usepackage[dvipsnames]{xcolor}
\newtheorem{theorem}{Theorem}

\newtheorem{lemma}{Lemma}
\newtheorem{corollary}{Corollary}
\newtheorem{rem}{Remark}

\DeclareMathOperator{\Tr}{Tr}
\DeclareMathOperator{\op}{op}
\DeclareMathOperator{\rank}{rank}
\DeclareMathOperator{\nuc}{\mathrm{nu}}
\DeclareMathOperator{\ii}{\bm{\mathrm{i}}} 
\DeclareMathOperator{\jj}{\bm{\mathrm{j}}} 
\DeclareMathOperator{\kk}{\bm{\mathrm{k}}} 
\DeclareMathOperator{\rr}{\bm{\mathrm{1}}}

\title{Color Image Inpainting via Robust Pure Quaternion Matrix Completion: Error Bound and Weighted Loss}
\author{Junren Chen\thanks{Department of 
		Mathematics, The University of Hong Kong, Pokfulam, Hong Kong (E-mail: chenjr58@connect.hku.hk).
		Thanks for the support from Hong Kong Research Grant Council HKPF scholarship.}
		\and Michael K. Ng\thanks{Department of 
		Mathematics, The University of Hong Kong, Pokfulam, Hong Kong (E-mail: mng@maths.hku.hk).
		Research supported in part by 
		Hong Kong Research Grant Council GRF 12300519, 17201020, 17300021.}
		}

\date{\today}
\begin{document}

\maketitle

% REQUIRED
\begin{abstract}
In this paper, we study color image inpainting as a pure quaternion matrix completion problem. In the literature, the theoretical guarantee for quaternion matrix completion is not well-established. Our main aim is to propose a new minimization problem with an objective combining nuclear norm and a quadratic loss weighted among three channels. To fill the theoretical vacancy, we obtain the error bound in both clean and corrupted regimes, which relies on some new results of quaternion matrices. A general Gaussian noise is considered in robust completion where all observations are corrupted. Motivated by the error bound, we propose to handle unbalanced or correlated noise via a cross-channel weight in the quadratic loss, with the main purpose of rebalancing noise level, or removing noise correlation. Extensive experimental results on synthetic and color image data are presented to confirm and demonstrate our theoretical findings.     
\end{abstract} 
% REQUIRED
%\begin{keywords}
 % example, \LaTeX
%\end{keywords}

% REQUIRED
%\begin{AMS}
 % 68Q25, 68R10, 68U05
%\end{AMS}

\section{Introduction}
\label{section1}
Compared with gray images, color images contain three highly correlated channels, i.e., R (red), G (green) and B (blue). A direct method in color image processing is the monochromatic model, which applies the well-developed gray image processing techniques to each channel separately, but usually results in performance degradation since it fails to capture the correlations among three channels \cite{chen2015color,chen2019low,xu2015vector,huang2021quaternion,sun2011color}. In addition, the concatenation model \cite{xu2015vector,xu2017multi} deals with the gray image obtained by unfolding and concatenating three channels, but it still cannot make full use of the couplings.

As an expansion of complex number, quaternion numbers contain one real part, three imaginary parts, and they form the non-commutative field $\mathbb{Q}$. We can completely capture the couplings among three channels by encoding RGB values into three imaginary parts of quaternion, then process color images as a whole under the non-trivial algebra structure of $\mathbb{Q}$. Therefore, $\mathbb{Q}$ has been noted to be a suitable platform for color image processing. 

The idea of using quaternion to represent and process color image can be traced back to \cite{sangwine1996fourier,pei1997novel,pei1999color}. After these initial applications, many classical tools in $\mathbb{R}$ or $\mathbb{C}$ were extended to $\mathbb{Q}$, to name a few, singular value decomposition \cite{zhang1997quaternions} and its applications \cite{chang2003quaternion,le2003color}, Fourier transform \cite{sangwine1996fourier,ell2006hypercomplex}, PCA \cite{sun2011color}, wavelet transform \cite{bayro2006theory}, moments analysis \cite{chen2012quaternion,chen2015color}. Moreover, some color image processing techniques based on quaternion were established, including denoising \cite{gai2015denoising,chen2019low,huang2021quaternion,yu2019quaternion,chen2014removing}, inpainting \cite{han2013color,mizoguchi2019hypercomplex,jia2019robust,miao2020quaternion}, segmentation \cite{shi2007quaternion,subakan2011quaternion}, watermarking \cite{bas2003color}, super-resolution \cite{yu2014single}, \textcolor{black}{deep neural network \cite{zhu2018quaternion,parcollet2019quaternion,le2021parameterized,gaudet2018deep,
parcollet2020survey,wu2020deep,quatpara}, graph embeddings \cite{quatgraph}, 
compressed sensing \cite{badenska2017compressed}, classification \cite{zeng2016color}, generative model \cite{grassucci2022quaternion} and many others.}

Thanks to the non-local self-similarity \cite{dabov2007image}, it is reasonable to assume natural images are (appriximately) low-rank, thus, color image inpainting can be cast as a low-rank quaternion matrix completion problem. Since a rank minimization problem is generally NP-hard \cite{gillis2011low}, we resort to rank surrogate that is computationally feasible. It has been proved that the nuclear norm is the tightest convex surrogate for the rank of quaternion matrices \cite{jia2019robust}, and undoubtedly nuclear norm is most frequently used in inpainting \cite{jia2019robust,mizoguchi2019hypercomplex,jia2020non,han2013color,miao2020low}. In pursuit of more precise approximation to rank, recent papers proposed to use non-convex surrogates that penalize large singular value less \cite{chen2019low,yu2019quaternion,yang2021low,yang2021weighted}, or some regularizers based on matrix factorization \cite{miao2020quaternion,miao2021color}, which avoids computing SVD of a large quaternion matrix but requires a prior estimation of the rank. Using quaternion to inpaint color image leads to exciting inpainting quality, we notice that more recent papers even began to study recovering color videos via quaternion tensor \cite{jia2020non,miao2020low}.

While existing work focused on algorithms, novel regularizers and experimental results, we notice that theoretical guarantees on reconstruction error are extremely rare. To our best knowledge, \cite{jia2019robust} is the only paper that provided exact recovery guarantee for low-rank quaternion matrix satisfying the incoherence condition \cite{candes2009exact,candes2010power,recht2011simpler}. However, it is impractical to assume natural images satisfy incoherence condition, which is stringent, unstable, and hard to verify. Without theoretical support on the recovery error, even the algorithms can perfectly find the global minimum point, the obtained quaternion matrix (i.e., reconstructed image) is not guaranteed to well approximate the underlying matrix (i.e., original image), hence quaternion-based inpainting method can only be an empirical success.

We fill the theoretical vacancy in this paper. We study the most classical quaternion inpainting method that uses nuclear norm as rank surrogate, and obtain the reconstructed error bound. Our results are established under much weaker conditions compared with \cite{jia2019robust}: the stringent incoherence condition is removed, and the original image is only assumed to be approximately low-rank, which is a relaxation of exact low-rankness. In regard to the robustness, we handle i.i.d. additive Gaussian noises distributed over the whole image, while \cite{jia2019robust} requires the noise to be sufficiently sparse. As a refinement, we restrict to pure quaternion matrix \cite{song2021low} with non-negative imaginary parts, for the RGB values of a pixel are non-negative. In comparison, most existing papers (e.g., \cite{miao2020low,jia2019robust,miao2021color}) neglect the "zero real part" constraint and choose to remove the real part of the reconstructed quaternion matrix, then the obtained pure quaternion matrix is not the best approximation, and more severely, it may not be low-rank, see Remark \ref{remark1}.

Moreover, we introduce a cross-channel weight in the loss function. When the noises in three channels have extremely different scales or are highly-correlated, we develop a noise correction strategy to rebalance the noise level, or remove the noise correlations. In brief, our main contributions can be summarized as follows:
\begin{itemize}
    \item We obtain the error bound of quaternion-based color image inpainting method that uses nuclear norm as rank surrogate. The error bound fills theoretical vacancy and manages to support existing literature. The established framework also encompasses the completion of a general quaternion matrix. 
    
    \item We consider general Gaussian noise in the corrupted model. To better achieve robustness under unbalanced or correlated noise, we propose a noise correction strategy that relies on a cross-channel weight in the loss function.  

\end{itemize}

\textcolor{black}{
The technical proof in our work is inspired by \cite{klopp2014noisy}, and can be viewed as a two-fold extension, i.e., to quaternion algebra and to approximately low-rank regime (note that \cite{klopp2014noisy} considered exactly low-rank matrix only). For extension to quaternion algebra, the main challenge is the lack of techincal tools available for quaternion matrices, e.g., even the fundamental relation in Lemma \ref{lemma2} is hard to position in literature. Thus, we develop some concentration and linear algebra results in Section 2 for the proof of the main results, and hopefully, they may facilitate quaternion matrices in the future research.
Compared with \cite{klopp2014noisy}, some steps in our proof are essentially different. For instance, the proof cannot be proceeded if we directly apply the Talagrand's contraction inequality after symmetrization. Instead, the components of quaternion matrix should be separated by triangle inequality before that, and to close the gap, one still needs to relate the nuclear norm of components to that of the quaternion matrix by Corollary \ref{corollary1}, see (\ref{B.7}) and (\ref{B.8}). Besides, 
the cross-channel weight in color imaging model based on 
quaternion numbers and general Gaussian noise add some complications to the proof.}

\textcolor{black}{Due to a modification of the constraint set (compare (30) in \cite{klopp2014noisy} and our (\ref{3.9})), our extension to approximately low-rankness seems quite new in the literature and hence may be interesting on its own manner. Indeed, the proof strategy in \cite{klopp2014noisy} was applied to establish error bound or statistical rate of different models in a bunch of follow-up papers, e.g., \cite{ding2017convex,lafond2014probabilistic,klopp2017robust,lafond2015low,klopp2015adaptive,athey2021matrix}, while these results are for exactly low-rank regime only. By using similar idea and modification in this work, it is possible to generalize them to approximately low-rank case. As a by-product, a much simpler proof for the main result in \cite{negahban2012restricted} can be obtained if Theorem 1 therein is substituted with a real version of our Lemma \ref{lemma5} (This can be found in Lemma 4 of our later work \cite{chen2022high}), see more discussions 
in Section 3.}

The paper is organized in this way: In Section \ref{section2} we provide preliminaries of quaternion and state the notations; In Section \ref{section3} we present our main results, including the reconstructed error bound under both clean and corrupted situations, and a noise correction strategy; In Section \ref{section4} we report experimental results on both synthetic and color image data to demonstrate and verify our theoretical findings; In Section \ref{section5} some concluding remarks are given. For clarity all proofs are collected in Appendix \ref{appendixA}, \ref{appendixB}. Due to space limit, the ADMM algorithms for pure quaternion matrix completion and additional experimental results on color images are provided in supplementary materials.

\section{Preliminaries and Notations}
\label{section2}

In principle, we use boldface lowercase letters (e.g., $\bm{\mathrm{q}}$, $\bm{\mathrm{\epsilon}}$) to denote quaternion numbers, while boldface capital letters (e.g., $\bm{X}$, $\bm{\Theta}$) represent matrices that can be real, complex or quaternionic. One exception is that we use $Y_1,...,Y_n$ to denote the $n$ observed values, which are quaternion basically.

Let $\mathbb{Q}$ be the set of quaternion, and $\bm{\mathrm{q}}=  q_0\rr + q_1\bm{\mathrm{i}} + q_2\jj +q_3\kk $ be a quaternion with $q_0\in \mathbb{R}$ as the real part (component), $q_1,q_2,q_3\in \mathbb{R}$ as three imaginary parts (components). The addition and subtraction of two quaternion numbers are operated component-wisely. By distributive law, associative law and the rule $\ii^2 = \jj^2 = \kk^2 = -1$, $\ii\jj = -\jj\ii = \kk$, $\jj \kk = -\kk \jj = \ii$, $\kk \ii = -\ii \kk = \jj$, it is straightforward to define multiplication. Note that the multiplication is non-commutative. We use $[.]_{\rr},[.]_{\ii},[.]_{\jj},[.]_{\kk}$ to extract the corresponding component of a quaternion number, vector or matrix, for example, $[\bm{\mathrm{q}}]_{\rr} = q_0$, $[\bm{\mathrm{q}}]_{\jj} = q_j$. Sometimes we also use the more standard notation $\mathrm{Re}(.)$ to extract the real part. The conjugate of $\bm{\mathrm{q}}$ is given by $\overline{\bm{\mathrm{q}}}= q_0\rr-  q_1\bm{\mathrm{i}} -q_2\jj -q_3\kk$, and we have $\bm{\mathrm{q}}\overline{\bm{\mathrm{q}}} = q_0^2+q_1^2+q_2^2+q_3^2 = |\bm{\mathrm{q}}|^2 ,$
where $|\bm{\mathrm{q}}|$ is called the absolute value or magnitude of $\bm{\mathrm{q}}$. The inverse of nonzero $\bm{\mathrm{q}}$ is $\bm{\mathrm{q}^{-1}} = \overline{\bm{\mathrm{q}}}/ |\bm{\mathrm{q}}|^2$, by which the division is defined immediately.

Consider a quaternion matrix $\bm{A}= [\bm{a_{ij}}],\bm{B}= [\bm{b_{ij}}] \in \mathbb{Q}^{M\times N}$, we let $\overline{\bm{A}} = [\bm{\overline{a_{ij}}}]$ be the conjugate, $\bm{A^T} = [\bm{a_{ji}}]\in \mathbb{Q}^{N\times M}$ be the transpose, and $\bm{A^*} = \bm{\overline{A}^T}= [\bm{\overline{a_{ji}}}]\in \mathbb{Q}^{N\times M}$ be the conjugate transpose. We define the standard inner product in $\mathbb{Q}^{M\times N}$ by 
$\big<\bm{A},\bm{B}\big> = \Tr(\bm{A^*B}) $, where $\Tr(.)$ denotes the trace of a matrix. \textcolor{black}{A set of quaternion vectors $\{\bm{V_1},...,\bm{V_N}\} \subset \mathbb{Q}^M$ is said to be right linearly independent if there exist no non-zero $(\bm{q_1},...,\bm{q_N})^T \in \mathbb{Q}^N \setminus \bm{0}$, such that $$\bm{V_1q_1}+\bm{V_2q_2}+...+\bm{V_Nq_N}=0.$$ Based on that}, the rank of $\bm{A}$ is defined to be the maximum number of right linearly independent columns vectors of $\bm{A}$. If $\bm{A}\in \mathbb{Q}^{N\times N}$ and $\rank(\bm{A}) = N$, then $A$ is invertible, and the inverse is denoted by $\bm{A^{-1}}$ satisfying $\bm{AA^{-1}} = \bm{A^{-1}A} = \bm{I_N}$. $\bm{A}$ is unitary if $\bm{A}$ is invertible and $\bm{A^{-1}}=\bm{A^*}$, $\bm{A}$ is Hermitian if $\bm{A^*} = \bm{A}$.

Quaternion singular value decomposition (QSVD) was introduced in \cite{zhang1997quaternions}. Given $\bm{A}\in \mathbb{Q}^{M\times N}$, there exist unitary matrices $\bm{U}\in \mathbb{Q}^{M\times M}$, $\bm{V}\in \mathbb{Q}^{N\times N}$, and diagonal matrix $\bm{\Sigma}= \mathrm{diag}(\sigma_1(\bm{A}),\sigma_2(\bm{A}),...,$   $\sigma_{\min\{M,N\}}(\bm{A}))\in \mathbb{Q}^{M\times N}$ such that
\begin{equation}
\bm{A} = \bm{U\Sigma V^*},
    \label{2.1}
\end{equation}
where non-negative numbers $\sigma_1(\bm{A})\geq \sigma_2(\bm{A}) \geq ... \geq \sigma_{\min\{M,N\}}(\bm{A})$ are the singular values of $\bm{A}$. Parallel to complex matrices, $\rank(\bm{A})$ equals the number of positive singular values of $\bm{A}$. We work with different matrix norms. The nuclear norm is defined to be the sum of singular values, i.e., $||\bm{A}||_{\nuc} = \sum_{i}\sigma_i(\bm{A})$. Besides, we define the Frobenius norm, operator norm, $||.||_{2,\infty}$ and max norm as ($[k]=\{1,...,k\}$)
\begin{equation}
\begin{aligned}
||\bm{A}||_{\mathrm{F}} = \Big[\sum_{i\in[M]}\sum_{j\in [N]}|\bm{a_{ij}}|^2\Big]^{1/2}\ &,\ ||\bm{A}||_{\op} = \sup_{\bm{x}\in \mathbb{Q}^N\setminus\{0\}}\frac{||\bm{Ax}||_2}{||\bm{x}||_2} \\
||\bm{A}||_{2,\infty} = \max_{i \in [M]} \Big[\sum_{j\in[N]} \bm{|\bm{a_{ij}}|^2} \Big]^{1/2}\ &,\ ||\bm{A}||_{\infty}= \max_{(i,j)\in [M]\times [N]}|\bm{a_{ij}}|,
\end{aligned}
    \nonumber
\end{equation}
where $||.||_2$ is the $\ell_2$ norm of a vector. %It is interesting to note that $$||\bm{A}||_{\mathrm{F}} = \Big[\sum_{i}\sigma_i(\bm{A})^2\Big]^{1/2}\ ,\ ||\bm{A}||_{\op} = \sigma_1(\bm{A}).$$

Another powerful tool to study quaternion matrices is the so-called complex adjoint matrix. Note that $\bm{A}\in \mathbb{Q}^{M\times N}$ can be uniquely written as $\bm{A^{(1)}}+ \bm{A^{(2)}}\jj$ where $\bm{A^{(1)}}$, $\bm{A^{(2)}}\in \mathbb{C}^{M\times N}$, then the complex adjoint matrix is given by
\begin{equation}
[\bm{A}]_{\mathbb{C}} = \begin{bmatrix} \bm{A^{(1)}} & \bm{A^{(2)}} \\ - \overline{\bm{A^{(2)}}} & \overline{\bm{A^{(1)}}}\end{bmatrix} \in \mathbb{C}^{2M\times 2N},
    \label{2.2}
\end{equation}
which preserves not only addition, subtraction, but also multiplication, conjugate transpose, inverse (Theorem 4.2, \cite{zhang1997quaternions}), e.g., $[\bm{A^*}]_{\mathbb{C}} = [\bm{A}]_{\mathbb{C}}^{\bm{*}}$, $[\bm{AC}]_{\mathbb{C}}= [\bm{A}]_{\mathbb{C}}[\bm{C}]_{\mathbb{C}}$ where $\bm{C}\in \mathbb{Q}^{N\times P}$. If necessary, $\bm{A}$ can be further reduced to a $4M\times 4N$ real matrix, see for example \cite{song2021low}.

A quaternion with zero real part is called pure quaternion, and the set of pure quaternion numbers is denoted by $\mathbb{Q}_{\mathrm{pure}}$ . Tailored to pixels of color image we consider pure quaternion with non-negative imaginary parts, which is called "pixel quaternion" and collected in the set $$\mathbb{Q}_{\mathrm{pix}}= \{ q_1\ii + q_2\jj + q_3\kk: q_1,q_2,q_3\in \mathbb{R}_{\geq 0} \}.$$
We mainly deal with pixel quaternion in this paper.

As a generalization of the quadratic loss, we consider a weighted loss where the weight $\bm{W}\in \mathbb{R}^{3\times 3}$ is a positive definite matrix, and $\bm{\sqrt{W}}$ stands for its square root. A pure quaternion $\bm{\mathrm{q}} = q_1\ii + q_2\jj+ q_3\kk$ is identified as 3-dimensional real vector $[q_1,q_2,q_3]^T\in \mathbb{R}^3$ when operating with $\bm{W}= [w_{ij}]$, for example, $$\begin{cases}\bm{\ \ W\mathrm{q}}&= \ (\sum_{i=1}^3w_{1i}q_i)\ii+(\sum_{i=1}^3w_{2i}q_i)\jj + (\sum_{i=1}^3w_{3i}q_i)\kk\\ \bm{\mathrm{q}^TW \mathrm{q}} &= \ \sum_{i=1}^3\sum_{j=1}^3q_iw_{ij}q_j\end{cases} \ .$$   
Moreover, we allow $\bm{W}$ to element-wisely operate on $\bm{A}= [\bm{a_{ij}}]\in \mathbb{Q}^{M\times N}$, i.e., $\bm{WA} = [\bm{Wa_{ij}}] \in  \mathbb{Q}^{M\times N}$. Since $\bm{W}$ is not restricted to be diagonal matrix, it introduces more flexibility in dealing with RGB correlations, see the noise correction strategy proposed in this work. For convenience we define the weighted magnitude of $\bm{\mathrm{q}} \in \mathbb{Q}_{\mathrm{pure}}$ to be $|\bm{\mathrm{q}}|_{\mathrm{w}} = \sqrt{\bm{\mathrm{q}}^T\bm{W}\bm{\mathrm{q}}}= |\bm{\sqrt{W}\mathrm{q}}|$, and the weighted Frobenius norm, weighted max norm are defined by $||\bm{A}||_{\mathrm{w},\mathrm{F}}= ||\bm{\sqrt{W}A}||_{\mathrm{F}},||\bm{A}||_{\mathrm{w},\infty}= ||\sqrt{\bm{W}}\bm{A} ||_{\infty}$, respectively.

The proof of our main results relies on some statistical methods. We use $\mathbbm{E}(.)$ and $\mathbbm{P}(.)$ to denote the expectation and the probability, respectively. Besides, if there exists an absolute constant $C$, such that $B_1 \leq C B_2$, then we write $B_1 \lesssim B_2$, we also use $B_1\gtrsim B_2$ to represent the converse inequality. We consider random Gaussian noise $\bm{\epsilon}\in \mathbb{Q}_{\mathrm{pure}}$ and assume that $([\bm{\epsilon}]_{\ii},[\bm{\epsilon}]_{\jj},[\bm{\epsilon}]_{\kk})^T\sim \mathcal{N}(0,\bm{\Sigma_{c}})$, where $\bm{\Sigma_c}\in \mathbb{R}^{3\times 3}$ is the positive definite covariance matrix.

To prove our main results we establish some technical tools for quaternion matrices, the proof can be found in Appendix \ref{appendixA}. Lemma \ref{lemma1}, Lemma \ref{lemma2} extend two well-known inequalities for complex matrices to quaternion matrices.
\begin{lemma}
Assume $\bm{A}\in \mathbb{Q}^{M\times N}$, then $||\bm{A}||_{\nuc}\leq \sqrt{\rank(\bm{A})}||\bm{A}||_{\mathrm{F}}$.
\label{lemma1}
\end{lemma}
\begin{lemma}
Assume $\bm{A,B}\in \mathbb{Q}^{M\times N}$, then $|\big<\bm{A,B}\big>|\leq ||\bm{A}||_{\op} ||\bm{B}||_{\nuc}$.
\label{lemma2}
\end{lemma}
 Lemma \ref{lemma3} shows the relationship between nuclear norm of quaternion matrix and its four components, while Corollary \ref{corollary1} is given in a generalized form. 
\begin{lemma}
Assume $\bm{A} = \bm{A_0} + \bm{A_{1}}\ii+ \bm{A_2}\jj + \bm{A_3}\kk\in \mathbb{Q}^{M\times N}$ where $\bm{A_0,A_1,A_2,A_3}\in \mathbb{R}^{M\times N}$, then we have $||\bm{A}||_{\nuc} \geq ||\bm{A_0}+ \bm{A_1}\ii||_{\nuc} \geq ||\bm{A_0} ||_{\nuc}$. 
\label{lemma3}
\end{lemma}
\begin{corollary}
Under the setting of Lemma \ref{lemma3}, we consider $\nu = (\nu_0,\nu_1,\nu_2,\nu_3)^T\in\mathbb{R}^4$, $|| \nu||_2 = 1$. Then we have $||\bm{A}||_{\nuc}\geq || \nu_0 \bm{A_0 } + \nu_1\bm{A_1}+ \nu_2\bm{A_2}+ \nu_3 \bm{A_3}||_{\nuc}$.
\label{corollary1}
\end{corollary}
\begin{rem}
Lemma \ref{lemma3} states that $|| \bm{A}||_{\nuc}\geq \max\{||\bm{A_0}+ \bm{A_1}\ii ||_{\nuc},||\bm{A_2}\jj+ \bm{A_3}\kk ||_{\nuc}\}$, which cannot be further improved to $|| \bm{A}||_{\nuc}\geq||\bm{A_1}\ii + \bm{A_2}\jj+ \bm{A_3}\kk ||_{\nuc}$. We consider quaternion matrix $\bm{A}$ and $\bm{A_p}$ obtained by removing the real part of $\bm{A}$: $$\bm{A} = \begin{bmatrix}1+3\ii & \kk - 3\jj \\ 1+ 3\jj & \kk + 3\ii \end{bmatrix} \ \ ;\ \bm{A_p} = \begin{bmatrix}3\ii & \kk - 3\jj \\  3\jj & \kk + 3\ii \end{bmatrix}.$$
It is not hard to verify $||\bm{A_p}||_{\nuc} > ||\bm{A}||_{\nuc}$. Moreover, $\rank(\bm{A})=1<\rank(\bm{A_p}) =2$, showing that removing the real part of a low-rank quaternion matrix may destory the low-rankness. Therefore, restriction to pure quaternion is important when dealing with color image inpainting.  
\label{remark1}
\end{rem}

Finally we provide the moment constraint version Bernstein inequality for quaternion matrices, which is an extension of Theorem 6.2 in \cite{tropp2012user}. 
%Although our main result (\ref{theorem2}) is presented under multivariate Gaussian noise, it directly extends to more general sub-exponential noise that satisfies the moment constraint \ref{2.3}. 
\begin{lemma}
Let $\bm{S_1}, \bm{S_2},...,\bm{S_n} \in \mathbb{Q}^{M\times M}$ be independent, Hermitian random quaternion matrices with zero mean, assume that for integer $p\geq 2$, there exist $R,\sigma>0$ such that  
\begin{equation}
||\mathbbm{E}\bm{S_k^p}||_{\op} \leq \frac{1}{2}p!R^{p-2}\sigma^2 .
    \label{2.3}
\end{equation}
Then for any $t>0$ we have 
\begin{equation}
\mathbbm{P}\Big[||\frac{1}{n}\sum_{k=1}^n\bm{S_k}||_{\op}\geq t\Big] \leq 4M\exp\left(-\frac{nt^2}{2(\sigma^2+ Rt)}\right) .
    \label{2.4}
\end{equation}
\label{lemma4}
\end{lemma}

\section{Main results}
\label{section3}
Let $\bm{\widetilde{\Theta}}= [\bm{\tilde{\theta}_{ij}}]\in \mathbb{Q}_{\mathrm{pix}}^{d_1\times d_2}$ be the original color image, $\bm{X_1},...,\bm{X_n}$ are i.i.d. uniformly distributed over $\{e_ie_j^T: i\in[d_1],j\in[d_2]\}$, or equivalently, assume $\bm{X_k}= e_{k(i)}e_{k(j)}^T$ where $(k(i),k(j))\sim \mathrm{unif}([d_1]\times [d_2])$. Note that $e_i$ is the vector whose i-th entry is 1 and other entries are 0, $e_i\in \mathbb{R}^{d_1}$, $e_j \in \mathbb{R}^{d_2}$, and so $e_ie_j^T\in \mathbb{R}^{d_1\times d_2}$, but we omit the dimension for simplicity. By using $\big<e_ie_j^T, \bm{\widetilde{\Theta}}\big> = \bm{\tilde{\theta}_{ij}}$ to represent observing the (i,j)-pixel, the clean model where the precise value of the entry is available can be formulated as
\begin{equation}
Y_k = \big<\bm{X_k,\widetilde{\Theta}} \big>.
    \label{3.1}
\end{equation}
We also study the robust completion problem where observations are corrupted by pure quaternion noise, i.e., 
\begin{equation}
Y_k = \big< \bm{X_k,\widetilde{\Theta}}\big> + \bm{\epsilon_k},
    \label{3.2}
\end{equation}
i.i.d. noise $\bm{\epsilon_k}$ are viewed as 3-dimensional real vectors and obey multivariate Gaussian distribution $\mathcal{N}(0,\bm{\Sigma_c})$. We study the problem of recovering $\bm{\widetilde{\Theta}}$ via $n$ incomplete samples $\{(\bm{X_k},Y_k): k\in[n]\}$ ($n<d_1d_2$).

Since $\bm{X_k}$ are i.i.d. distributed, one position may be observed twice or even more in our model, while existing work (e.g., \cite{jia2019robust,mizoguchi2019hypercomplex,miao2020quaternion}) considered a seemingly quite different sampling scheme without replacement, i.e., one entry is either observed once or unknown. To relate them, we point out that under the same sample size, sampling without replacement is at least as good as sampling with replacement \cite{foygel2011concentration}, and this is obvious for the clean model (\ref{3.1}), since observing one position twice doesn't provide more information. In a nutshell, we are indeed coping with more difficult sampling scheme, hence our results manage to support the problem settings of existing papers.

\textcolor{black}{
For our main results, there are two obvious extensions that can be obtained by a bit more efforts but no technical difficulty. Firstly, the random sampling scheme need not to be uniform, instead it can have a rather general probability distribution over the image, see \cite{klopp2014noisy}. Thus, the random sampling scheme captures many real situations such as damaged pixels in old photos, images exposed to bad conditions, and image compression based on low-rank structure \cite{nguyen2019low}. Secondly, the assumption of Gaussian noise is inessential, and actually the result extends to correlated sub-exponential noise, since sub-exponential tail suffices to admit (\ref{2.3}) in Lemma \ref{lemma4}. Random contamination due to intrinsic thermal and electronic fluctuations of the acquisition devices is known to be the main source of noise \cite{luisier2010image}. Although it is conventional to consider additive white Gaussian noise, sub-exponential noise undoubtedly model it more precisely, as the exact noise pattern is also related to the sensor setting, see for example \cite{liu2007automatic}. 
In addition, the statistics of color noise is not independent of RGB channels because of the demosaic process embedded in a camera \cite{liu2007automatic}, which accounts for our consideration of noise correlations.}

\textcolor{black}{
For succinctness, we present our main results under uniform random sampling and Gaussian noise, and leave these simple extensions to avid readers.}

We first consider the clean case without noise (\ref{3.1}). Our method requires a prior estimation of the max norm 
\begin{equation}
|| \bm{\widetilde{\Theta}} ||_{\infty} \leq \alpha\ .
    \label{3.3}
\end{equation}
Since the gray value is smaller than 255, $\alpha$ can always be $255\sqrt{3}$, but a tighter bound on $|| \bm{\widetilde{\Theta}} ||_{\infty}$ would be preferable and bring higher inpainting quality. The precise observations are used as constraints, by combining with (\ref{3.3}) we complete the color image via minimizing the nuclear norm,
\begin{equation}
\bm{\widehat{\Theta}} \in \mathop{\arg\min}\limits_{\bm{\Theta} \in \mathbb{Q}_{\mathrm{pix}}^{d_1\times d_2}}  \ || \bm{\Theta}||_{\nuc}, \ \mathrm{s.t.}\ ||\bm{\Theta}||_{\infty} \leq \alpha,\ \big<\bm{X_k},\bm{\Theta}\big> = Y_k,\ k\in [n].
    \label{3.4}
\end{equation}

In (\ref{3.2}) the observations are corrupted and not completely reliable, so we use a weighted quadratic loss to fit the observed data: 
\begin{equation}
\mathcal{L}_{\mathrm{w}}(\bm{\Theta}) = \frac{1}{2n}\sum_{k=1}^n | Y_{k} - \big<\bm{X_k,\Theta}\big>|_{\mathrm{w}}^2\ ,\ \bm{\Theta} \in \mathbb{Q}_{\mathrm{pix}}^{d_1\times d_2}.
    \label{3.5}
\end{equation}
To avoid scaling confusion we assume $\Tr \bm{W} = ||\bm{W} ||_{\nuc} = \sum_i \lambda_i(\bm{W}) =3$, which means assigning a total weight $3$ to three channels for diagonal $\bm{W}$, and note that when $\bm{W} = \bm{I_3}$ (\ref{3.5}) reduces to the commonly used quadratic loss \cite{yu2019quaternion,mizoguchi2019hypercomplex,chen2019low,miao2021color,huang2021quaternion}.
In accordance to the weight $\bm{W}$ we require a prior estimation on the weighted max norm
\begin{equation}
||\bm{\widetilde{\Theta}} ||_{\mathrm{w},\infty} \leq \alpha \ . 
    \label{3.6}
\end{equation}
Let the nuclear norm serve as regularizer that promote low-rankness, and $\lambda$ be the tuning parameter to balance the loss and regularization, the inpaint method is formulated as 
\begin{equation}
\bm{\widehat{\Theta}} \in \mathop{\arg\min}\limits_{\bm{\Theta} \in\mathbb{Q}_{\mathrm{pix}}^{d_1\times d_2}}  \  \mathcal{L}_{\mathrm{w}}(\bm{\Theta})+ \lambda ||\bm{\Theta}||_{\nuc},\ \mathrm{s.t.}\ ||\bm{\Theta} ||_{\mathrm{w},\infty} \leq \alpha.
    \label{3.7}
\end{equation}
Although natural image is believed to contain high redundancies, it need not to be exactly low-rank. To be more practical, we assume $\bm{\widetilde{\Theta}}$ to be approximately low-rank under a specific $0\leq q <1$:
\begin{equation}
\sum_{i=1}^{\min\{d_1,d_2\}} |\sigma_i(\bm{\widetilde{\Theta}})|^q\leq \rho\ , 
    \label{3.8}
\end{equation}
which reduces to low-rank when $q=0$.

\subsection{Error Bound}
Let $\bm{\widehat{\Delta}} = \bm{\widehat{\Theta}- \widetilde{\Theta}}$ be the reconstructed error, in this part we bound $||\bm{\widehat{\Delta}}||_{\mathrm{F}}$ for (\ref{3.1}), (\ref{3.2}). Different from matrix sensing that has an elegant theory parallel to compressed sensing \cite{cai2013sparse}, it is challenging to establish restricted isometry property (RIP) without incoherence condition in matrix completion \cite{davenport2016overview}. To replace RIP, \cite{negahban2012restricted} first established the restricted strong convexity (RSC) over a constraint set and obtained an error bound, but the quite lengthy and intricate proof involves construction of $\delta$-net, Sudakov minoration, hence may not be easy to follow. By considering a different constraint set, \cite{klopp2014noisy} successfully refined the proof for exactly low-rank regime. 

\textcolor{black}{
Similarly, the RSC of $\mathcal{L}_{\mathrm{w}}(\bm{\Theta})$ presented in Lemma \ref{lemma5} is a key ingredient for our main result. It should be stressed that Lemma \ref{lemma5} two-fold extends Lemma 12 in \cite{klopp2014noisy} to quaternion algebra and approximately low-rank regime. The first extension is due to some new handling for concentration or inequality of quaternion matrices, while the second one owes to a tricky modification of the critical constraint set, specifically, the constraint set $\mathcal{C}(r)$ in \cite{klopp2014noisy} is adjusted to $\mathcal{C}(\psi)$ in (\ref{3.9}), which depends on the parameter $q$ in (\ref{3.8}). Note that the second extension is of independent technical interest: We already point out that it indicates possible extensions of many existing results in literature; And since \cite{klopp2014noisy} is a refinement of exactly low-rank case of \cite{negahban2012restricted}, our proof provides an alternative and much simpler proof for the main result in \cite{negahban2012restricted}. }

\begin{lemma}
Consider the constraint set with parameter $\psi = (\psi_1,\psi_2,\psi_3)$ where $\psi_3$ is slightly large \begin{equation}
\begin{aligned}
\mathcal{C}(\psi) = \Big\{ \bm{\Theta}\in \mathbb{Q}_{\mathrm{pure}}^{d_1\times d_2}: &||\bm{\Theta}||_{\mathrm{w},\infty}\leq \psi_1\alpha,||\bm{\Theta}||_{\nuc} \leq \psi_2\rho^{\frac{1}{2-q}}|| \bm{\Theta}||_{\mathrm{F}}^{\frac{2-2q}{2-q}} , \\ 
&|| \bm{\Theta} ||_{\mathrm{w},\mathrm{F}}^2\geq \alpha^2 d_1d_2\sqrt{\frac{\psi_3\log(d_1+d_2)}{n}} \Big\} .
\end{aligned}
    \label{3.9}
\end{equation}
Let $\mathscr{X} = (\bm{X_1},...,\bm{X_n})$, $\bm{\Theta} \in \mathbb{Q}_{\mathrm{pure}}^{d_1\times d_2}$, we define 
\begin{equation}
\mathcal{F}_{\mathscr{X}}(\bm{\Theta}) = \frac{1}{n}\sum_{k=1}^n |\big<\bm{X_k,\Theta} \big>|^2_{\mathrm{w}}\ .
    \label{3.10}
\end{equation}
Then there exists constant $\kappa \in (0,1)$ such that with probability at least $1-2(d_1+d_2)^{-3/4}$, for all $\bm{\Theta}  \in\mathcal{C}(\psi)$ it holds that 
\begin{equation}
\mathcal{F}_{\mathscr{X}}(\bm{\Theta}) \geq \frac{\kappa}{d_1d_2}|| \bm{\Theta}||_{\mathrm{w},\mathrm{F}}^2 - T_0,
    \label{3.11}
\end{equation}
where the relaxation term $T_0$ is given by
\begin{equation}
T_0 = \frac{\rho}{(2-q)(\sqrt{d_1d_2})^q} \left(36\psi_1\psi_2\alpha \frac{|| \bm{W}||_{\infty}^{1/2}}{\lambda_{\min}(\bm{\sqrt{W}})}\sqrt{\frac{\max\{d_1,d_2\}\log (d_1+d_2)}{n}}  \right)^{2-q}. 
    \label{3.12}
\end{equation}
\label{lemma5}
\end{lemma}

\vspace{2mm}
The core idea of the proof is to discuss whether the error $\bm{\widehat{\Delta}}$ falls in $\mathcal{C}(\psi)$, which helps unfold some key relations that further lead to an error bound. By (\ref{3.3}), (\ref{3.6}), it is straightforward that $\bm{\widehat{\Delta}}$ satisfies the first constraint in (\ref{3.9}). In Lemma \ref{lemma6} we show that in both clean and corrupted cases, the error satisfies the second constraint for certain $\psi_2$. The proof relies on analysis of a pair of subspace $(\mathcal{M},\overline{\mathcal{M}})$ over which the nuclear norm is decomposable, see the unified framework established in \cite{negahban2012unified}.  

\begin{lemma}
Assume (\ref{3.8}) holds.

\noindent
(1) In the clean model (\ref{3.1}) with prior estimation (\ref{3.3}), we recover $\bm{\widetilde{\Theta}}$ via (\ref{3.4}), then we have \begin{equation}
 || \bm{\widehat{\Delta}} ||_{\nuc} \leq 5\rho^{\frac{1}{2-q}}|| \bm{\widehat{\Delta}}||_{\mathrm{F}}^{\frac{2-2q}{2-q}}.
    \label{3.13}
\end{equation}

\noindent
(2) In the corrupted model (\ref{3.2}) with prior estimation (\ref{3.6}), we recover $\bm{\widetilde{\Theta}}$ via (\ref{3.7}) with the weighted loss in (\ref{3.5}). If for some $C_1>1$, $\lambda$ satisfies
\begin{equation}
 \lambda \geq C_1 ||\frac{1}{n} \sum_{k=1}^n (\bm{W\epsilon_k}) \bm{X_k} ||_{\op}
    \label{3.14}
\end{equation}then we have 
\begin{equation}
|| \bm{\widehat{\Delta}} ||_{\nuc} \leq \frac{5C_1}{C_1-1}\rho^{\frac{1}{2-q}}|| \bm{\widehat{\Delta}}||_{\mathrm{F}}^{\frac{2-2q}{2-q}}. 
    \label{3.15}
\end{equation}
\label{lemma6}
\end{lemma}
With the aid of Lemma \ref{lemma5}, Lemma \ref{lemma6}, we are now in a position to derive error bound of the clean model (\ref{3.1}). We give an upper bound on the mean square error (MSE), i.e., $||\bm{\widehat{\Delta}} ||_{\mathrm{F}}^2 /(d_1d_2)$, which directly links to the commonly used peak signal-to-noise ratio (PSNR), or the Frobenius norm error $|| \bm{\widehat{\Delta}}||_{\mathrm{F}}$.

\begin{theorem}
Consider the clean model (\ref{3.1}) with prior estimation (\ref{3.3}), the color image is reconstructed by (\ref{3.4}). Assume the original image satisfies (\ref{3.8}). If $n<d_1d_2$, $\rho \gtrsim (d_1d_2)^{q/2}$, $\max\{d_1,d_2\}\log(d_1+d_2)/n$ is sufficiently small, the error bound for MSE holds 
\begin{equation}
\frac{|| \bm{\widehat{\Delta}}||_{\mathrm{F}}^2}{d_1d_2} \lesssim \frac{\rho}{(d_1d_2)^{q/2}}\left( \alpha^2 \frac{\max\{d_1,d_2\}\log(d_1+d_2)}{n}\right)^{1-q/2}
    \label{3.16}
\end{equation}
with probability at least $1-2(d_1+d_2)^{-3/4}$.
\label{theorem1}
\end{theorem}
\begin{rem}
From (\ref{3.16}), sampling complexity $n=O(\max\{d_1,d_2\}\log(d_1+d_2))$ suffices to guarantee a small MSE, but it may not be surprising since MSE is a more element-wise notion. To be more insightful, we rewrite (\ref{3.16}) to be a bound of relative square error (RSE): 
\begin{equation}
    \frac{||\bm{\widehat{\Delta}} ||_{\mathrm{F}}^2}{||\bm{\widetilde{\Theta}} ||_{\mathrm{F}}^2} \lesssim  \frac{\rho}{||\bm{\widetilde{\Theta}} ||_{\mathrm{F}}^q} \left(\alpha(\bm{\widetilde{\Theta}})^2\frac{\max\{d_1,d_2\}\log(d_1+d_2)}{n}\right)^{1-q/2}, 
    \label{3.17}
\end{equation}
where $\alpha(\bm{\widetilde{\Theta}})$ given by  $$\alpha(\bm{\widetilde{\Theta}})= \frac{\sqrt{d_1d_2}||\bm{\widetilde{\Theta}} ||_{\infty}}{||\bm{\widetilde{\Theta}} ||_{\mathrm{F}}}\in [1,\sqrt{d_1d_2}]$$ is known as the spikiness of $\bm{\widetilde{\Theta}}$, which was first proposed in \cite{negahban2012restricted} for a real matrix. Therefore, a small RSE can be guaranteed under $n = O(\max\{d_1,d_2\}\log(d_1+d_2))$ for $\bm{\widetilde{\Theta}}$ with spikiness scaling as a constant. Note that natural (color) images tend to be smooth locally \cite{blomgren1998color} and contains similar patches globally \cite{dabov2007image}, so they are supposed to have low spikiness. This can also be experimentally verified, and we report that most color images tested in our numerical simulation admit spikiness lower than 2. In comparison, the exact recovery framework in \cite{jia2019robust} cannot fit natural images well due to the unrealistic "incoherence" assumption,  which is more stringent than low spikiness \cite{negahban2012restricted}.

\label{remark2}
\end{rem}

For the corrupted case we need Lemma \ref{lemma7} to give instruction on tuning the $\lambda$, which is expected to satisfy (\ref{3.14}) so that (\ref{3.15}) holds. After all these preparations, in Theorem \ref{theorem2} we establish an error bound under general Gaussian noise and a general weighted loss, by similar methodology.  

\begin{lemma}
In (\ref{3.7}) if we choose \begin{equation}
\lambda = 2C_1\sqrt{\frac{\Tr(\bm{W\Sigma_{c} W})\log(d_1+d_2)}{n\min\{d_1,d_2\}}},
    \label{3.18}
\end{equation}
then (\ref{3.14}) holds with probability higher than $1-4(d_1+d_2)^{-3/4}$. 
\label{lemma7}
\end{lemma}

\begin{theorem}
Consider the corrupted model (\ref{3.2}) where $\bm{\epsilon_k}\sim \mathcal{N}(0,\bm{\Sigma_c})$, given the positive definite weight $\bm{W}$ satisfying $\Tr\bm{W}=3$, we have the prior estimation (\ref{3.6}). Assume the original image satisfies (\ref{3.8}). We reconstruct the color image via (\ref{3.7}) where the loss function is given in (\ref{3.5}). We choose $\lambda $ by (\ref{3.18}). If $n<d_1d_2$, $\rho \gtrsim (d_1d_2)^{q/2}$, $\max\{d_1,d_2\}\log(d_1+d_2)/n$ is small, then with probability at least $1-6(d_1+d_2)^{-3/4}$ we have the error bound for MSE
\begin{equation}
\frac{||\bm{\widehat{\Delta}} ||_{\mathrm{F}}^2}{d_1d_2} \lesssim \frac{\rho}{(d_1d_2)^{q/2}}\left(\max\Big\{\frac{||\bm{W} ||_{\infty}\alpha^2}{\lambda_{\min}(\bm{W})^3},\frac{\Tr(\bm{W\Sigma_c\bm{W}})}{\lambda_{\min}(\bm{W})^2} \Big\}\frac{\max\{d_1,d_2\}\log(d_1+d_2)}{n}\right)^{1-q/2}.
    \label{3.19}
\end{equation}
\label{theorem2}
\end{theorem}

\subsection{Noise Correction}
%Except for the intrinsic correlations, cross-channel noise may generate more mix-ups among three channels. 
 Due to the i.i.d. Gaussian noise that corrupts each observation, we are required to conduct completion and denoising simultaneously when we cope with the robust completion problem. How to model and then handle cross-channel noise has been noted to be an important issue in color image denoising \cite{nam2016holistic,xu2017multi}, thus, a careful treatment for the noise is also necessary in our robust inpainting problem (\ref{3.2}). Motivated by the obtained error bound (\ref{3.19}), we propose to use a cross-channel weight $\bm{W}$ to handle unbalanced or highly correlated noise, which is referred to be a noise correction strategy.

Compared with the bound of clean model (\ref{3.16}), the bound under corrupted observations contains two extra factors involving the noise covariance matrix $\bm{\Sigma_c}$ and the cross-channel weight $\bm{W}$, i.e., 
$$F_1 = \frac{||\bm{W}||_{\infty}}{\lambda_{\min}(\bm{W})^3}\ \ \mathrm{and} \ \ F_2 = \frac{\Tr(\bm{W\Sigma_cW})}{\lambda_{\min}(\bm{W})^2}. $$
Under assumption $\Tr(\bm{W})=3$, it can be easily verified that $||\bm{W}||_{\infty}\geq 1$, $\lambda_{\min}(\bm{W})\leq 1$, for both inequalities the equal sign holds if and only if $\bm{W} = \bm{I_3}$, meaning that the error bound becomes worse when $\bm{W}$ deviates from $\bm{I_3}$. The guideline that $\bm{W}$ should not be far away from the identity matrix is mainly because the non-weighted loss (i.e., $\bm{W}= \bm{I_3}$) best fits the square error (or MSE, RSE, PSNR) that we care about. On the other hand, the numerator of $F_2$, $\Tr(\bm{W}\bm{\Sigma_c}\bm{W})$, seems to indicate an appropriate weight for the noise.

\subsubsection{Noise Level Rebalance}
In some real applications prior knowledge on noise levels is available, also some efficient methods to estimate noise variance in image data have been developed \cite{liu2013single,chen2015efficient}. We assume the diagonal of $\bm{\Sigma_c}$ is known to be $[\sigma_r^2,\sigma_g^2,\sigma_b^2]^T$. In the corrupted model, a noise with remarkably different $\sigma_r^2,\sigma_g^2,\sigma_b^2$ could be problematic, since the observations in three channels have different data fidelities. More precisely, data in the channel suffering from severer noise will be less reliable than those in low noise channel.

To deal with the issue of data fidelity, we use a diagonal $\bm{W} = \mathrm{diag}(w_r,w_g,w_b)$ that assign higher weight to channel with more precise data, to rebalance the noise level. It is not hard to show that \begin{equation}
    \bm{W_s}  = \frac{3\mathrm{diag}(\sigma_r^{-2},\sigma_g^{-2},\sigma_b^{-2})}{\sigma_r^{-2}+\sigma_g^{-2}+\sigma_{b}^{-2}}
    \label{3.20}
\end{equation} is the only diagonal matrix minimizing $\Tr(\bm{W\Sigma_c W})$. Except for the motivation from the bound, such a choice can also be justified by a simple reasoning. Note that (\ref{3.2}) is equivalent to 
\begin{equation}
\bm{\sqrt{W_s}}Y_k = \big<\bm{X_k},\bm{\sqrt{W_s}}\bm{\widetilde{\Theta}} \big> +\bm{\sqrt{W_s}}\bm{\epsilon_k},
    \label{3.21}
\end{equation}
and $\bm{\sqrt{W_s}}\bm{\epsilon_k}$ have equal noise levels, so data in three channels of $\bm{\sqrt{W_s}}\bm{\widetilde{\Theta}}$ have equal data fidelity, therefore, a fair loss would be 
\begin{equation}
    \begin{aligned}
    \frac{1}{2n}\sum_{k=1}^n |\bm{\sqrt{W_s}}Y_k - \big<\bm{X_k},\bm{\sqrt{W_s}\Theta}\big>|^2  
    =  \frac{1}{2n}\sum_{k=1}^n | \bm{\sqrt{W_s}}(Y_k - \big< \bm{X_k},\bm{\Theta}\big>)|^2 
    =  \mathcal{L}_{\mathrm{w}_s}(\bm{\Theta}).
    \nonumber
    \end{aligned}
\end{equation}

In summary, on one hand, $\bm{W}$ cannot deviate much from $\bm{I_3}$ to maintain the alignment of the loss $\mathcal{L}_{w}(\bm{\Theta})$ and the considered square error $||\bm{\widehat{\Delta}} ||^2_{\mathrm{F}}$; on the other hand, $\bm{W}$ close to $\bm{W_s}$ is preferred thanks to its ability to rebalance noise levels. As a trade-off, our proposal is to use the convex combination
\begin{equation}
\bm{W}(\gamma) = (1-\gamma)\bm{I_3}+ \gamma \bm{W_s},
    \label{3.22}
\end{equation}
where $\gamma \in [0,1].$

We mention that the idea of handling unbalanced noise via weighting three channels was first proposed in Jun Xu et al. \cite{xu2017multi}, but it worked on denoising which is different from our robust inpainting problem. More importantly, they failed to see the requirement that the weight cannot go far away from the identity matrix, and set the weight to be $\bm{W}(1) = \bm{W_s}$. We will use experimental results to show that a proper $\gamma$ in (\ref{3.22}) is crucial, while the extreme choice $\bm{W}(1)=\bm{W_s}$ seems inadvisable, and could even lead to degraded result worse than a non-weighted loss $\bm{W}(0)=\bm{I_3}.$

\subsubsection{Noise Correlation Removal}
Although noise of different scales has been considered in \cite{xu2017multi}, few work studied correlated noise in color images, e.g., \cite{xu2017multi} assumed noises in three channels are independent. In this part, we report that highly-correlated noise could also be detrimental, probably because such noise introduces extra cross-channel correlations that mixes with the original RGB couplings, making it more difficult to recognize the underlying color image. Then, we propose to use a positive definite weight $\bm{W}$ to remove (some) correlations of the noise. To focus on the strategy per se, we assume we have full knowledge of $\bm{\Sigma_c}$, and select a $\bm{W}$ that minimize $\Tr(\bm{W\Sigma_cW})$.

\begin{lemma}
Under the assumption $\Tr\bm{W}=3$, and $\bm{W}$ is positive definite. Then \begin{equation}
    \bm{W_c}=\frac{3\bm{\Sigma_c}^{-1}}{\Tr(\bm{\Sigma_c}^{-1})}
    \label{3.23}
\end{equation} is the unique weight that minimizes $\Tr(\bm{W\Sigma_c W})$.
\label{lemma8}
\end{lemma}
In analogy, the advantage of (\ref{3.23}) can also be seen by transforming the model as (\ref{3.21}), and note that $\bm{\sqrt{W_c}\epsilon_k}\sim \mathcal{N}(0,3\bm{I_3}/\Tr(\bm{\Sigma_c}^{-1}))$. As a trade-off among $\bm{I_3}$, $\bm{W_s}$ and $\bm{W_c}$, we propose to choose $\bm{W}$ by \begin{equation}
\bm{W}(\gamma_1,\gamma_2) = (1-\gamma_1-\gamma_2) \bm{I_3} + \gamma_1 \bm{W_s} + \gamma_2 \bm{W_c},
     \label{3.24}
\end{equation}
where $\gamma_1>0$, $\gamma_2>0$, $\gamma_1+\gamma_2\leq 1$. When the noise correlations are benign, or $\bm{\Sigma_c}$ is unknown, we simply let $\gamma_2 = 0$, then (\ref{3.24}) reduces to our previous proposal (\ref{3.22}).

\section{Experimental results}
\label{section4}

\subsection{Synthetic Data}
In this part we use synthetic (approximately) low-rank non-negative pure quaternion matrix (NNPQM) for numerical simulations, aiming to illustrate the relations in the obtained error bound and verify the effectiveness of the proposed noise correction strategy. Also we compare sampling with or without replacement. We apply ADMM algorithms \cite{boyd2011distributed} with guaranteed convergence \cite{gabay1976dual,he20121} to solve the proposed convex optimization problem (\ref{3.4}), (\ref{3.7}), see supplementary materials for the details.

\begin{figure}[ht]
    \centering
    \includegraphics[scale = 0.65]{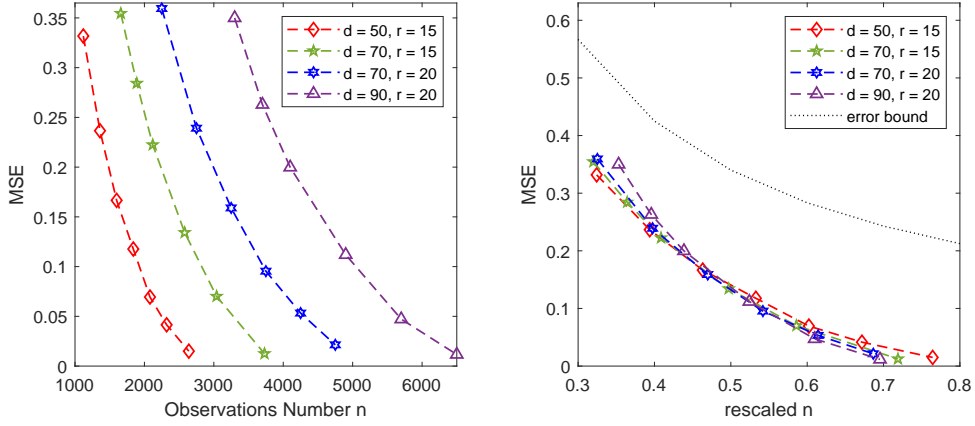}
    \caption{\sl q = 0, MSE v.s. n or $\frac{n}{rd\log(2d)}$ (rescaled n) under different d, r}
    \label{fig1}
\end{figure}

\begin{figure}[ht]
    \centering
    \includegraphics[scale = 0.75]{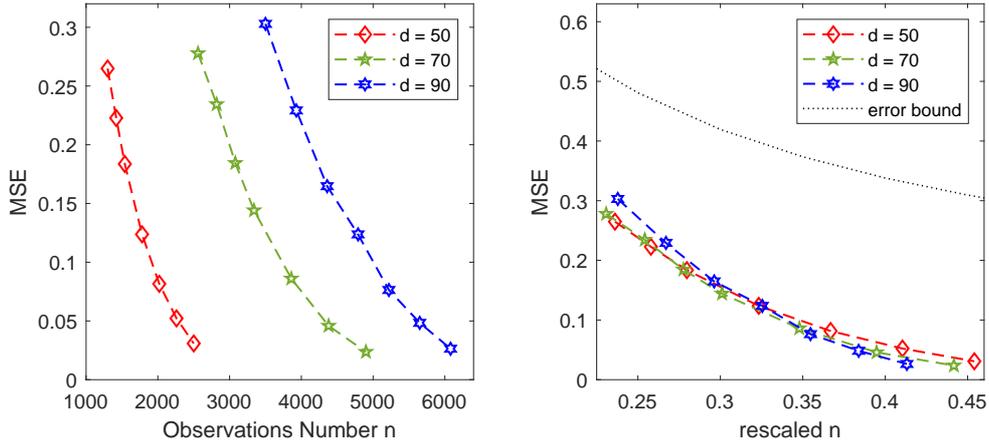}
    \caption{\sl q = $\frac{1}{2}$, MSE v.s. n or $\frac{n}{(\rho d^{-q})^{2/(2-q)}d\log(2d)}$ (rescaled n) under different d, $\rho$}
    \label{fig2}
\end{figure}

\begin{figure}[ht]
    \centering
    \includegraphics[scale = 0.85]{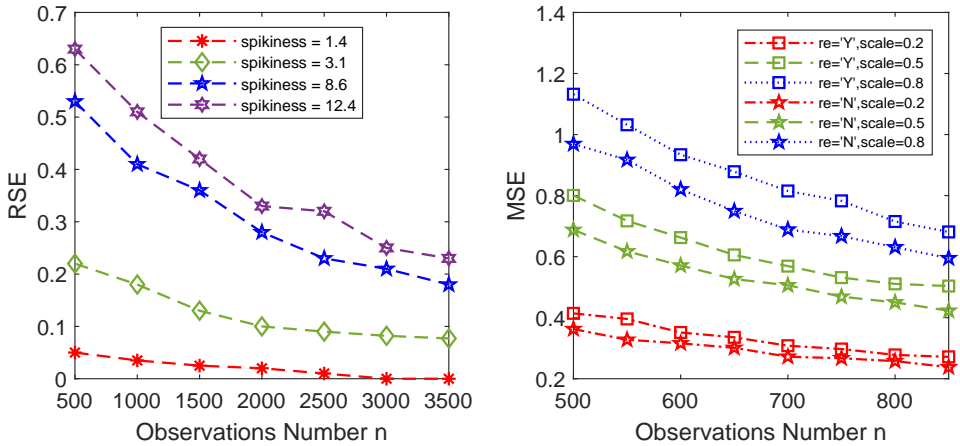}
    \caption{ \sl Left figire: RSE v.s. n of NNPQM with the same size, rank, but different spikiness values; Right Figure: MSE v.s. n under different noise levels (scales), re = 'Y' or 'N' stands for sampling with or without replacement, respectively.}
    \label{fig4}
\end{figure}

\textcolor{black}{
Assume $\bm{L_i},\bm{Q},\bm{Q_i}$ are random matrices with i.i.d. entries drawn from uniform distribution over $[0,1]$. We use the following random mechanism to generate $d_1\times d_2$ NNPQM with rank r: Generate random matrices $\bm{L_0},\bm{L_1},\bm{L_2} \in \mathbb{R}^{d_1\times (r-1)}$, find a basis of the null space of $[\bm{L_0},\bm{L_1},\bm{L_0}+\bm{L_2}, \bm{L_1}-\bm{L_2}]$, and then collect them as columns in $\bm{K}\in \mathbb{R}^{4(r-1)\times d_3}$; Use random matrix $\bm{Q}\in \mathbb{R}^{d_3\times d_2}$ to obtain $ \bm{K}\bm{Q}=[\bm{R_0}^T,\bm{R_1}^T,\bm{R_2}^T,\bm{R_3}^T]^T \in \mathbb{R}^{4(r-1)\times d_2}$ with blocks $\bm{R_l}\in \mathbb{R}^{(r-1)\times d_2}$; Construct quaternion matrices $\bm{L} = \bm{L_0} - \bm{L_1}\ii  - (\bm{L_0}+\bm{L_2})\jj - (\bm{L_1}-\bm{L_2})\kk \in \mathbb{Q}^{d_1\times (r-1)}$, $\bm{R} = \bm{R_0}+ \bm{R_1}\ii + \bm{R_2}\jj + \bm{R_3}\kk \in \mathbb{Q}^{(r-1)\times d_2}$, then we obtain $\bm{LR}\in \mathbb{Q}^{d_1\times d_2}$ with zero real part and $\mathrm{rank}(\bm{LR}) = r-1$; Finally we apply a "rank-1 lifting" to render non-negativity (let $[.]$ be the rounding function, e.g., $[-2.5] = -3$, $\mathbf{1}_{d_1,d_2}$ be the $d_1\times d_2$ all-ones matrix)
$$ \bm{ \widetilde{\Theta} }= \bm{LR} -\left( \Big[\min [\bm{LR}]_{\ii} \Big]\ii -\Big[\min [\bm{LR}]_{\jj} \Big]\jj-\Big[\min [\bm{LR}]_{\kk} \Big]\kk \right) \mathbf{1}_{d_1,d_2}, $$
and this is the rank r NNPQM we use. For approximately low-rank we just randomly perturb an exactly low-rank NNPQM $\bm{\widetilde{\Theta}_0}$, specifically we set $\tau = \min \big[[\bm{\widetilde{\Theta}_0}]_{\ii},[\bm{\widetilde{\Theta}_0}]_{\jj},[\bm{\widetilde{\Theta}_0}]_{\kk}\big]$ and use
$$\bm{\widetilde{\Theta}} =\bm{\widetilde{\Theta}_0} - 0.1*\tau*\Big[ \bm{Q_1}\ii+ \bm{Q_2} \jj + \bm{Q_3}\kk\Big]. $$
A uniform sampling scheme with replacement is used in experiments in Figure \ref{fig1}, Figure \ref{fig2}, Figure \ref{fig4}, while uniform sampling without replacement is adopted for the rest.}

For exact low-rank NNPQM (\ref{3.16}) reads 
\begin{equation}
    \mathrm{MSE}\lesssim \alpha^2 \mathrm{rank}(\bm{\widetilde{\Theta}}) \frac{\max\{d_1,d_2\}\log(d_1+d_2)}{n},
    \label{4.3}
\end{equation}
To confirm the relations among MSE, n, rank and dimension of underlying matrix, we generate square NNPQMs with different rank r and dimension d, and use the same $\alpha$. MSE for each matrix under a specific n is set as the mean value of 10 repetitions. The plots of "MSE v.s. n" for synthetic NNPQM with "r = 50, d = 15", "r = 70, d = 15", "r = 70, d = 20", "r = 90, d = 20" are showed in Figure \ref{fig1}. We can see that the curves in the left figure shift to the right with larger d or r, since more observations are needed to complete a quaternion matrix of larger size or higher rank. Define the rescaled sample size $n_{\mathrm{re}} =n/(rd\log 2d) $, as showed in the right figure, several curves of "MSE v.s. $n_{\mathrm{re}}$" are fairly aligned, showing that MSE is directly proportional to "$rd\log 2d/n$" quite precisely. \textcolor{black}{In addition, we provide the curve of theoretical bound $\frac{c}{n_{\mathrm{re}}}$ for comparison (we choose $c = 0.17$ here). Note that from the viewpoint of convergence rate, $O(\frac{1}{n_{\mathrm{re}}})$ was showed to be minimax optimal in \cite{negahban2012restricted}.}

The approximate low-rank case under (\ref{3.8}) with $q=\frac{1}{2}$ is also numerically simulated. We generate NNPQM of size $50\times 50$, $70\times 70$, $90\times90$, all are  of full rank but only a few singular values are significant. Define the rescaled sample size $n_{\mathrm{re}} = n/ \rho^{4/3}d^{1/3}\log (2d)$, the plots of "MSE v.s. n" and "MSE v.s. $n_{\mathrm{re}}$" showed in Figure \ref{fig2} are consistent with the obtained error bound (\ref{3.16}). \textcolor{black}{The curve of theoretical bound $\frac{c}{n^{3/4}_{\mathrm{re}}}$ with $c = 0.15$ is also provided.}

In Remark \ref{remark2} the obtained error bound indicates that the relative square error (RSE) is also related to the spikiness of the underlying NNPQM, more precisely, it is harder to complete a more spiky matrix. This relation is quite evident under the most spiky NNPQM that has only one nonzero entry, since such a matrix cannot be well estimated unless nearly all entries are observed \cite{davenport2016overview}. To further verify that such a relation between RSE and spikiness is not proof artifact but a general fact, we generate $50\times 50$ rank-20 NNPQMs with different spikiness values, then obtain the RSE that is set to be the mean value of 15 repetitions. The results are showed on the left side of Figure \ref{fig4}, as expected, the NNPQM with lower spikiness can be better completed under the same conditions. Besides, we found that most real color images have spikiness lower than 2 (see Remark \ref{remark2} for underlying rationale), thus, our framework nicely embraces color images. 

%\begin{figure}[h]
 %   \centering
  %  \includegraphics[scale = 0.54]{new_spikiness.eps}
   % \caption{\sl RSE v.s. n, under different spikiness, the same d, r}
%    \label{fig3}
%\end{figure}

It has been showed in \cite{foygel2011concentration} that no-replacement sampling is at least as good as sampling with replacement used in (\ref{3.1}), (\ref{3.2}). As a double check we compare two sampling schemes on a fixed $30\times 30$ rank-5 NNPQM corrupted by Gaussian noise of different levels. From the results on the right side of Figure \ref{fig4}, sampling without replacement seems more informative and leads to smaller error under the same conditions. In the following experiments, we adopt sampling without replacement since it is more commonly used in literature.

\begin{figure}[ht]
    \centering
    \includegraphics[scale =0.95]{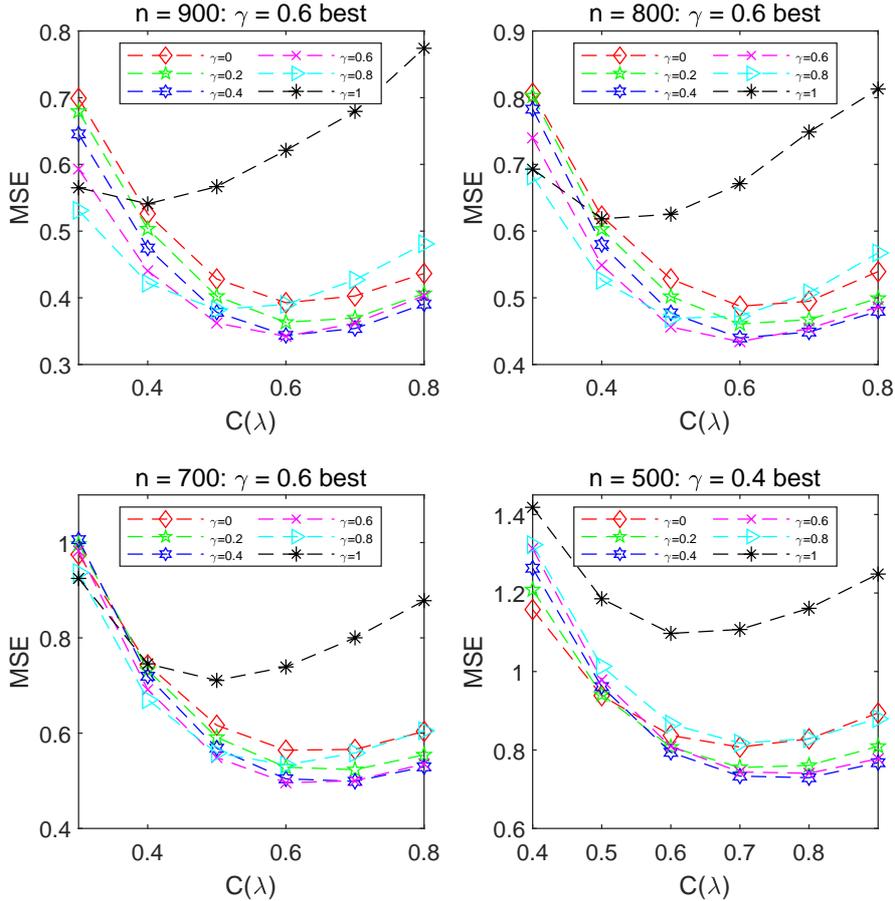}
    \caption{\sl(Noise level rebalance) MSE v.s. $C(\lambda)$ for $\gamma$ = 0, 0.2, 0.4, 0.6, 0.8, 1, under n = 900, 800, 700, 500. The best $\gamma$ corresponding to the smallest MSE is given in the title.}
    \label{fig5}
\end{figure}

Recall that we propose to deal with unbalanced or correlated noise via choosing a suitable cross-channel weight, including noise level rebalance and noise correlation removal, see (\ref{3.22}), (\ref{3.24}) for the precise proposals. We first confirm the efficacy of "rebalancing noise level" on a synthetic $30\times 30$ rank-5 NNPQM, whose entries are corrupted by i.i.d. unbalanced Gaussian noise with covariance matrix $\bm{\Sigma_c}=\mathrm{diag}(1.5,0.5,0.2)$. By (\ref{3.20}) the weight that completely achieves equal data fidelity is $\bm{W_s} = \mathrm{diag(2/9,10/9,5/3)}$, so we use $\bm{W}(\gamma) = \mathrm{diag}(1-7\gamma/9,1+\gamma/9,1+2\gamma/3)$ as a trade-off. In the experiment we examined $\gamma$ = 0, 0.2, 0.4, 0.6, 0.8, 1.0, where $\gamma =0$ becomes the non-weighted case, and $\gamma = 1$ corresponds to the proposal in \cite{xu2017multi}. Note that a proper scaling of $\lambda$ in (\ref{3.7}) is provided in (\ref{3.18}), thus, we set
\begin{equation}
\lambda = C(\lambda)\sqrt{\frac{\Tr(\bm{W}\bm{\Sigma_c}\bm{W})\log(2d)}{nd}}
    \label{4.4}
\end{equation}
for specific $\bm{W}$. $C(\lambda)$ is tuned to be (nearly) optimal for a fair comparison among different $\bm{W}(\gamma)$, and in this simulation the optimal $C(\lambda)$ always lies between 0.4 and 0.8. In Figure \ref{fig5} we show the results under sample size 900, 800, 700, 500, when "n = 900" the robust completion problem becomes a denoising problem (recall that we sample without replacement).

From Figure \ref{fig5}, we see that a properly weighted loss brings significant improvement: When n equals 900 or 800, the curves of $\bm{W}(0.2)$, $\bm{W}(0.4)$, $\bm{W}(0.6)$ uniformly lie below that of $\bm{W}(0)$; When n equals 700, 500, the minimum MSE of $\bm{W}(0.4)$ or $\bm{W}(0.6)$ is still notably smaller compared with the non-weighted loss. It is interesting to see that a full balance of noise level, i.e., $\bm{W}(1)=\bm{W_s}$, could result in severe degradation.

In analogy we show removing noise correlation is an effective strategy to deal with correlated noise. We generate a $30\times 30$ rank-2 NNPQM and then corrupt its entries by i.i.d. Gaussian noise, which is highly, positively correlated with covariance matrix
$\bm{\Sigma_c}=[0.70,0.50,0.50;$ $0.50,0.70,0.66;0.50,0.66,0.70].$
Note that noise levels in three channels are the same, hence $\bm{W_s}=\bm{I_3}$, and (\ref{3.24}) reduces to \begin{equation}
    \bm{W}(\gamma)= (1-\gamma)\bm{I_3}+\gamma \bm{W_c},
    \label{4.5}
\end{equation}
where $\bm{W_c}$ is determined by (\ref{3.23}). Similarly we test $\gamma$ = 0, 0.2, 0.4, 0.6, 0.8, 1.0 under $n$ = 900 (denoising), 700, 400, 200. $\lambda$ is set as (\ref{4.4}) where $C(\lambda)$ is well tuned.

\begin{figure}[ht!]
    \centering
    \includegraphics[scale = 0.9]{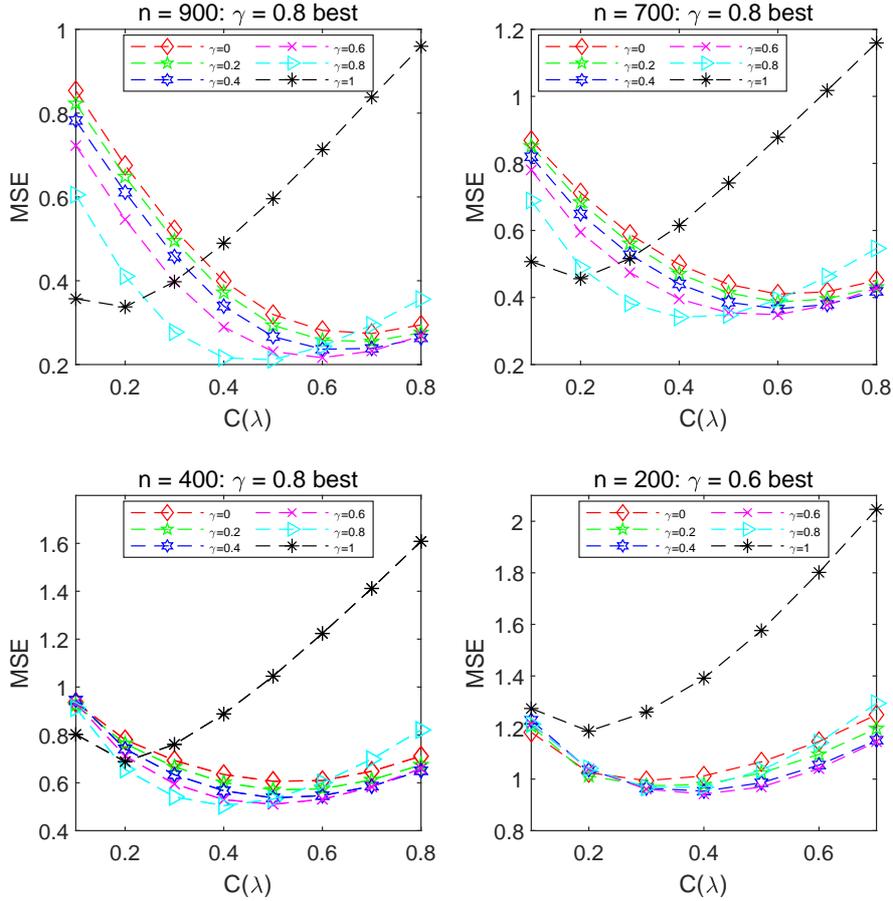}
    \caption{\sl(Noise correlation removal) MSE v.s. $C(\lambda)$ for $\gamma$ = 0, 0.2, 0.4, 0.6, 0.8, 1, under n = 900, 700, 400, 200. The best $\gamma$ corresponding to the smallest MSE is given in the title.}
    \label{fig6}
\end{figure}

From Figure \ref{fig6} we see that In denoising (n=900) and robust completion where n = 700 or 400, $\bm{W}(0.2)$, $\bm{W}(0.4)$, $\bm{W}(0.6)$ are uniformly better than a non-weighted loss, while $\bm{W}(0.8)$ achieves the smallest square error. Under a smaller sample size n = 200, the minimum MSE achieved by "$\gamma$ = 0.6" notably improves that of "$\gamma$ = 0". In addition, using $\bm{W}(1)=\bm{W_c}$ that removes all correlations of noise leads to result worse than non-weighted loss, since it deviates from the identity matrix too much.

\begin{rem}\label{remark3}
This remark is intended to provide an insight of $\gamma$. Since each observation is corrupted by noise in (\ref{3.2}), the robust completion problem can be viewed as a combination of completion and denoising. The only purpose of using $\bm{W_s}$ or $\bm{W_c}$ in $\bm{W}(\gamma)$ is to adjust to the noise pattern and facilitate the denoising. By contrast, the other component $\bm{I_3}$ aims to achieve the alignment between loss function and square error, hence it is crucial for the whole problem. From this perspective, $\gamma$ is actually weighting "to what extent the robust completion is a denoising problem". It is interesting to note that in Figure \ref{fig5}, Figure \ref{fig6} when n becomes smaller, the optimal $\gamma$ moves towards 0, and the curves of different $\gamma$ become closer, indicating that the improvement is less significant. This is because the completion dominates the whole problem while denoising plays a relatively minor role under small sample size.
\end{rem}

\subsection{Color Image}

In this section we switch to real color image data. As mentioned above, the main advantage of quaternion-based inpainting method is that it completely captures the correlations among three channels. Under clean model we compare it with two real counterparts, i.e., completing each channel separately, or completing the concatenation matrix, see the details and results in supplementary materials. Interested readers may refer to \cite{jia2019robust} for comparisons between different types of methods (e.g., tensor-based method).

\begin{figure}[ht]
    \centering
    \includegraphics[scale = 0.57]{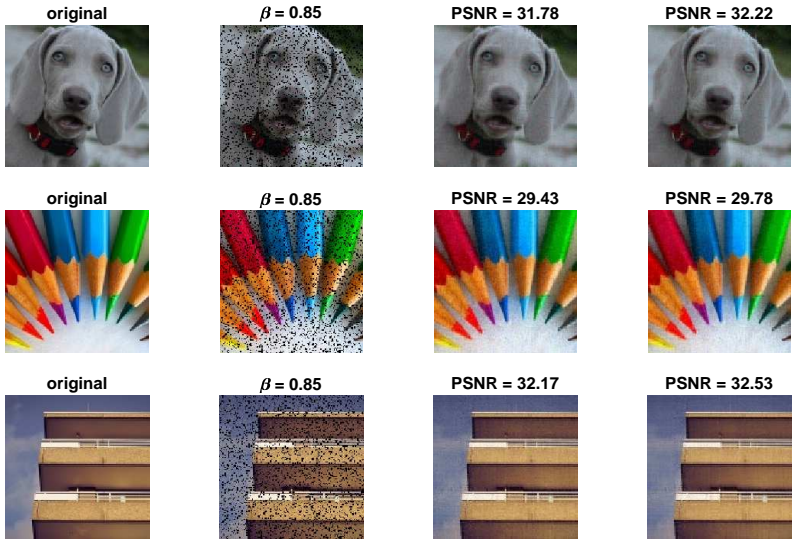} 
    \caption{\sl (Correcting the noise in real data) The first column: original images; The second column: observed images with 85\% of the entries; The third column: images reconstructed by non-weighted loss; The last column: images reconstructed by properly weighted loss.} 
    \label{fig9}
\end{figure}

The noise correction proposal is one of our main contributions. To present its successful application in real data, we provide three numerical examples on $128\times 128$ color images from the Linnaeus 5 dataset. More additional experiments can be found in the supplementary materials. Let $ \mathcal{G}=[0,0.1,0.2,0.3,0.4,0.5,0.6,0.7,0.8,0.9,1.0]$. In the first row of Figure \ref{fig9} we use an unbalanced noise with $\bm{\Sigma_c}= \mathrm{diag}(90,10,15)$, hence the R channel suffers from much severer noise, then we apply a loss weighted by $\bm{W}(\gamma)$ in (\ref{3.22}) with $\gamma$ tuned over $\mathcal{G}$. $\bm{W}(0.7)$ was found to be optimal, see the result in the last column. In the second row the noise covariance matrix is $\bm{\Sigma_c}=[75,70,65;70,75,63;65,63,75]$, (\ref{3.24}) (now reduced to $\bm{W}(\gamma) = (1-\gamma)\bm{I_3}+\gamma \bm{W_c}$) is used to remove some noise correlations. We tuned $\gamma$ over $\mathcal{G}$ and found $\bm{W}(0.5)$ is the best. The noise in the third row has covariance matrix given by $\bm{\Sigma_c}=[75,70,0;70,75,0;0,0,5]$, thus, it is highly correlated in R, G channels, and also unbalanced since the noise level in channel R, G is much higher than channel B. Based on (\ref{3.24}), $\bm{W}(\gamma_1,\gamma_2)= (1-\gamma_1-\gamma_2)\bm{I_3}+\gamma_1\bm{W_s}+\gamma_2\bm{W_c}$ is employed to cope with the noise, where $(\gamma_1,\gamma_2)$ is tuned over $\mathcal{G}\times \mathcal{G}$. The result under the optimal choice $\bm{W}(0.7,0.1)$ is given in the last column. The improvement on PSNR report the success of our proposal to adapt to the noise pattern.

$$~$$

\vspace{8mm}

\section{Concluding Remarks}
\label{section5}

Quaternion-based method for color image inpainting was widely studied in literature. We notice that most works focus on algorithm design and convergence, or aim to propose new rank surrogate to replace nuclear norm, while the theoretical guarantee on reconstructed error remains under-developed. To our best knowledge, \cite{jia2019robust} is the only paper that presents an exact recovery guarantee, however, the stringent incoherence condition and exact low-rankness are required, which may not be satisfied by color images.

The main aim of this paper is to fill the theoretical vacancy. In this work, we formulate color image inpainting as a robust pure quaternion matrix completion problem, and obtain an error bound under much more relaxed conditions that nicely match color images. Moreover, we achieve stronger robustness than previous work \cite{jia2019robust} since all observations are corrupted, while the corrupted entries are assumed to be sufficiently sparse in their work.

To be more general, we consider a quadratic loss weighted among three channels (\ref{3.5}), and the obtained error bound motivates us to handle unbalanced or highly-correlated noise via a suitable cross-channel weight. More specifically, we propose two noise correction strategies with the core spirit of rebalancing noise level, or removing noise correlation. It is interesting to note that \cite{xu2017multi} also used a diagonal weight in color image denoising, the weight therein corresponds to $\bm{W}(1)$ in (\ref{3.22}). However, we use numerical results Figure \ref{fig5} to report that such an extreme choice may not be sensible. The underlying rationale is also provided, see our analysis on factors $F_1$, $F_2$.

Our main results are restricted to completing a non-negative pure quaternion matrix because of the background of color image inpainting, however, all of them directly extend to general quaternion matrix completion, which is also an under-studied topic. In completion of a general quaternion matrix from corrupted observations, for example, a quadratic loss weighted by a suitable $4\times 4$ positive definite matrix can be applied to adapt to the noise pattern, which achieves stronger robustness.   

%\bibliographystyle{siamplain}
%\bibliography{references}

\begin{thebibliography}{10}
 	
 	\bibitem{athey2021matrix}
 	{\sc S.~Athey, M.~Bayati, N.~Doudchenko, G.~Imbens, and K.~Khosravi}, {\em
 		Matrix completion methods for causal panel data models}, Journal of the
 	American Statistical Association, 116 (2021), pp.~1716--1730.
 	
 	\bibitem{badenska2017compressed}
 	{\sc A.~Bade{\'n}ska and {\L}.~B{\l}aszczyk}, {\em Compressed sensing for real
 		measurements of quaternion signals}, Journal of the Franklin Institute, 354
 	(2017), pp.~5753--5769.
 	
 	\bibitem{bas2003color}
 	{\sc P.~Bas, N.~Le~Bihan, and J.-M. Chassery}, {\em Color image watermarking
 		using quaternion fourier transform}, in 2003 IEEE International Conference on
 	Acoustics, Speech, and Signal Processing, 2003. Proceedings.(ICASSP'03).,
 	vol.~3, IEEE, 2003, pp.~III--521.
 	
 	\bibitem{bayro2006theory}
 	{\sc E.~Bayro-Corrochano}, {\em The theory and use of the quaternion wavelet
 		transform}, Journal of Mathematical Imaging and Vision, 24 (2006),
 	pp.~19--35.
 	
 	\bibitem{blomgren1998color}
 	{\sc P.~Blomgren and T.~F. Chan}, {\em Color tv: total variation methods for
 		restoration of vector-valued images}, IEEE transactions on image processing,
 	7 (1998), pp.~304--309.
 	
 	\bibitem{boyd2011distributed}
 	{\sc S.~Boyd, N.~Parikh, and E.~Chu}, {\em Distributed optimization and
 		statistical learning via the alternating direction method of multipliers},
 	Now Publishers Inc, 2011.
 	
 	\bibitem{cai2013sparse}
 	{\sc T.~T. Cai and A.~Zhang}, {\em Sparse representation of a polytope and
 		recovery of sparse signals and low-rank matrices}, IEEE transactions on
 	information theory, 60 (2013), pp.~122--132.
 	
 	\bibitem{candes2009exact}
 	{\sc E.~J. Cand{\`e}s and B.~Recht}, {\em Exact matrix completion via convex
 		optimization}, Foundations of Computational mathematics, 9 (2009),
 	pp.~717--772.
 	
 	\bibitem{candes2010power}
 	{\sc E.~J. Cand{\`e}s and T.~Tao}, {\em The power of convex relaxation:
 		Near-optimal matrix completion}, IEEE Transactions on Information Theory, 56
 	(2010), pp.~2053--2080.
 	
 	\bibitem{chang2003quaternion}
 	{\sc J.-H. Chang, J.-J. Ding, et~al.}, {\em Quaternion matrix singular value
 		decomposition and its applications for color image processing}, in
 	Proceedings 2003 International Conference on Image Processing (Cat. No.
 	03CH37429), vol.~1, IEEE, 2003, pp.~I--805.
 	
 	\bibitem{chen2014removing}
 	{\sc B.~Chen, Q.~Liu, X.~Sun, X.~Li, and H.~Shu}, {\em Removing gaussian noise
 		for colour images by quaternion representation and optimisation of weights in
 		non-local means filter}, IET Image Processing, 8 (2014), pp.~591--600.
 	
 	\bibitem{chen2015color}
 	{\sc B.~Chen, H.~Shu, G.~Coatrieux, G.~Chen, X.~Sun, and J.~L. Coatrieux}, {\em
 		Color image analysis by quaternion-type moments}, Journal of mathematical
 	imaging and vision, 51 (2015), pp.~124--144.
 	
 	\bibitem{chen2012quaternion}
 	{\sc B.~Chen, H.~Shu, H.~Zhang, G.~Chen, C.~Toumoulin, J.-L. Dillenseger, and
 		L.~Luo}, {\em Quaternion zernike moments and their invariants for color image
 		analysis and object recognition}, Signal processing, 92 (2012), pp.~308--318.
 	
 	\bibitem{chen2015efficient}
 	{\sc G.~Chen, F.~Zhu, and P.~Ann~Heng}, {\em An efficient statistical method
 		for image noise level estimation}, in Proceedings of the IEEE International
 	Conference on Computer Vision, 2015, pp.~477--485.
 	
 	\bibitem{chen2022high}
 	{\sc J.~Chen, C.-L. Wang, M.~K. Ng, and D.~Wang}, {\em High dimensional
 		statistical estimation under one-bit quantization}, arXiv preprint
 	arXiv:2202.13157,  (2022).
 	
 	\bibitem{chen2019low}
 	{\sc Y.~Chen, X.~Xiao, and Y.~Zhou}, {\em Low-rank quaternion approximation for
 		color image processing}, IEEE Transactions on Image Processing, 29 (2019),
 	pp.~1426--1439.
 	
 	\bibitem{dabov2007image}
 	{\sc K.~Dabov, A.~Foi, V.~Katkovnik, and K.~Egiazarian}, {\em Image denoising
 		by sparse 3-d transform-domain collaborative filtering}, IEEE Transactions on
 	image processing, 16 (2007), pp.~2080--2095.
 	
 	\bibitem{davenport2016overview}
 	{\sc M.~A. Davenport and J.~Romberg}, {\em An overview of low-rank matrix
 		recovery from incomplete observations}, IEEE Journal of Selected Topics in
 	Signal Processing, 10 (2016), pp.~608--622.
 	
 	\bibitem{ding2017convex}
 	{\sc C.~Ding and H.-D. Qi}, {\em Convex optimization learning of faithful
 		euclidean distance representations in nonlinear dimensionality reduction},
 	Mathematical Programming, 164 (2017), pp.~341--381.
 	
 	\bibitem{ell2006hypercomplex}
 	{\sc T.~A. Ell and S.~J. Sangwine}, {\em Hypercomplex fourier transforms of
 		color images}, IEEE Transactions on image processing, 16 (2006), pp.~22--35.
 	
 	\bibitem{foygel2011concentration}
 	{\sc R.~Foygel and N.~Srebro}, {\em Concentration-based guarantees for low-rank
 		matrix reconstruction}, in Proceedings of the 24th Annual Conference on
 	Learning Theory, JMLR Workshop and Conference Proceedings, 2011,
 	pp.~315--340.
 	
 	\bibitem{gabay1976dual}
 	{\sc D.~Gabay and B.~Mercier}, {\em A dual algorithm for the solution of
 		nonlinear variational problems via finite element approximation}, Computers
 	\& mathematics with applications, 2 (1976), pp.~17--40.
 	
 	\bibitem{gai2015denoising}
 	{\sc S.~Gai, G.~Yang, M.~Wan, and L.~Wang}, {\em Denoising color images by
 		reduced quaternion matrix singular value decomposition}, Multidimensional
 	Systems and Signal Processing, 26 (2015), pp.~307--320.
 	
 	\bibitem{gaudet2018deep}
 	{\sc C.~J. Gaudet and A.~S. Maida}, {\em Deep quaternion networks}, in 2018
 	International Joint Conference on Neural Networks (IJCNN), IEEE, 2018,
 	pp.~1--8.
 	
 	\bibitem{gillis2011low}
 	{\sc N.~Gillis and F.~Glineur}, {\em Low-rank matrix approximation with weights
 		or missing data is np-hard}, SIAM Journal on Matrix Analysis and
 	Applications, 32 (2011), pp.~1149--1165.
 	
 	\bibitem{grassucci2022quaternion}
 	{\sc E.~Grassucci, E.~Cicero, and D.~Comminiello}, {\em Quaternion generative
 		adversarial networks}, in Generative Adversarial Learning: Architectures and
 	Applications, Springer, 2022, pp.~57--86.
 	
 	\bibitem{han2013color}
 	{\sc X.~Han, J.~Wu, L.~Yan, L.~Senhadji, and H.~Shu}, {\em Color image recovery
 		via quaternion matrix completion}, in 2013 6th International Congress on
 	Image and Signal Processing (CISP), vol.~1, IEEE, 2013, pp.~358--362.
 	
 	\bibitem{he20121}
 	{\sc B.~He and X.~Yuan}, {\em On the o(1/n) convergence rate of the
 		douglas--rachford alternating direction method}, SIAM Journal on Numerical
 	Analysis, 50 (2012), pp.~700--709.
 	
 	\bibitem{huang2021quaternion}
 	{\sc C.~Huang, M.~K. Ng, T.~Wu, and T.~Zeng}, {\em Quaternion-based dictionary
 		learning and saturation-value total variation regularization for color image
 		restoration}, IEEE Transactions on Multimedia,  (2021).
 	
 	\bibitem{jia2020non}
 	{\sc Z.~Jia, Q.~Jin, M.~K. Ng, and X.~Zhao}, {\em Non-local robust quaternion
 		matrix completion for large-scale color images and videos inpainting}, arXiv
 	preprint arXiv:2011.08675,  (2020).
 	
 	\bibitem{jia2019robust}
 	{\sc Z.~Jia, M.~K. Ng, and G.-J. Song}, {\em Robust quaternion matrix
 		completion with applications to image inpainting}, Numerical Linear Algebra
 	with Applications, 26 (2019), p.~e2245.
 	
 	\bibitem{klopp2014noisy}
 	{\sc O.~Klopp}, {\em Noisy low-rank matrix completion with general sampling
 		distribution}, Bernoulli, 20 (2014), pp.~282--303.
 	
 	\bibitem{klopp2015adaptive}
 	{\sc O.~Klopp, J.~Lafond, {\'E}.~Moulines, and J.~Salmon}, {\em Adaptive
 		multinomial matrix completion}, Electronic Journal of Statistics, 9 (2015),
 	pp.~2950--2975.
 	
 	\bibitem{klopp2017robust}
 	{\sc O.~Klopp, K.~Lounici, and A.~B. Tsybakov}, {\em Robust matrix completion},
 	Probability Theory and Related Fields, 169 (2017), pp.~523--564.
 	
 	\bibitem{lafond2015low}
 	{\sc J.~Lafond}, {\em Low rank matrix completion with exponential family
 		noise}, in Conference on Learning Theory, PMLR, 2015, pp.~1224--1243.
 	
 	\bibitem{lafond2014probabilistic}
 	{\sc J.~Lafond, O.~Klopp, E.~Moulines, and J.~Salmon}, {\em Probabilistic
 		low-rank matrix completion on finite alphabets}, Advances in Neural
 	Information Processing Systems, 27 (2014).
 	
 	\bibitem{le2021parameterized}
 	{\sc T.~Le, M.~Bertolini, F.~No{\'e}, and D.-A. Clevert}, {\em Parameterized
 		hypercomplex graph neural networks for graph classification}, in
 	International Conference on Artificial Neural Networks, Springer, 2021,
 	pp.~204--216.
 	
 	\bibitem{le2003color}
 	{\sc N.~Le~Bihan and S.~J. Sangwine}, {\em Color image decomposition using
 		quaternion singular value decomposition}, in 2003 International Conference on
 	Visual Information Engineering VIE 2003, IET, 2003, pp.~113--116.
 	
 	\bibitem{liu2007automatic}
 	{\sc C.~Liu, R.~Szeliski, S.~B. Kang, C.~L. Zitnick, and W.~T. Freeman}, {\em
 		Automatic estimation and removal of noise from a single image}, IEEE
 	transactions on pattern analysis and machine intelligence, 30 (2007),
 	pp.~299--314.
 	
 	\bibitem{liu2013single}
 	{\sc X.~Liu, M.~Tanaka, and M.~Okutomi}, {\em Single-image noise level
 		estimation for blind denoising}, IEEE transactions on image processing, 22
 	(2013), pp.~5226--5237.
 	
 	\bibitem{luisier2010image}
 	{\sc F.~Luisier, T.~Blu, and M.~Unser}, {\em Image denoising in mixed
 		poisson--gaussian noise}, IEEE Transactions on image processing, 20 (2010),
 	pp.~696--708.
 	
 	\bibitem{miao2020quaternion}
 	{\sc J.~Miao and K.~I. Kou}, {\em Quaternion-based bilinear factor matrix norm
 		minimization for color image inpainting}, IEEE Transactions on Signal
 	Processing, 68 (2020), pp.~5617--5631.
 	
 	\bibitem{miao2021color}
 	{\sc J.~Miao and K.~I. Kou}, {\em Color image recovery using low-rank
 		quaternion matrix completion algorithm}, IEEE Transactions on Image
 	Processing,  (2021).
 	
 	\bibitem{miao2020low}
 	{\sc J.~Miao, K.~I. Kou, and W.~Liu}, {\em Low-rank quaternion tensor
 		completion for recovering color videos and images}, Pattern Recognition, 107
 	(2020), p.~107505.
 	
 	\bibitem{mizoguchi2019hypercomplex}
 	{\sc T.~Mizoguchi and I.~Yamada}, {\em Hypercomplex low rank matrix completion
 		with non-negative constraints via convex optimization}, in ICASSP 2019-2019
 	IEEE International Conference on Acoustics, Speech and Signal Processing
 	(ICASSP), IEEE, 2019, pp.~8538--8542.
 	
 	\bibitem{nam2016holistic}
 	{\sc S.~Nam, Y.~Hwang, Y.~Matsushita, and S.~J. Kim}, {\em A holistic approach
 		to cross-channel image noise modeling and its application to image
 		denoising}, in Proceedings of the IEEE conference on computer vision and
 	pattern recognition, 2016, pp.~1683--1691.
 	
 	\bibitem{negahban2012restricted}
 	{\sc S.~Negahban and M.~J. Wainwright}, {\em Restricted strong convexity and
 		weighted matrix completion: Optimal bounds with noise}, The Journal of
 	Machine Learning Research, 13 (2012), pp.~1665--1697.
 	
 	\bibitem{negahban2012unified}
 	{\sc S.~N. Negahban, P.~Ravikumar, M.~J. Wainwright, and B.~Yu}, {\em A unified
 		framework for high-dimensional analysis of $ m $-estimators with decomposable
 		regularizers}, Statistical science, 27 (2012), pp.~538--557.
 	
 	\bibitem{nguyen2019low}
 	{\sc L.~T. Nguyen, J.~Kim, and B.~Shim}, {\em Low-rank matrix completion: A
 		contemporary survey}, IEEE Access, 7 (2019), pp.~94215--94237.
 	
 	\bibitem{parcollet2019quaternion}
 	{\sc T.~Parcollet, M.~Morchid, and G.~Linar{\`e}s}, {\em Quaternion
 		convolutional neural networks for heterogeneous image processing}, in ICASSP
 	2019-2019 IEEE International Conference on Acoustics, Speech and Signal
 	Processing (ICASSP), IEEE, 2019, pp.~8514--8518.
 	
 	\bibitem{parcollet2020survey}
 	{\sc T.~Parcollet, M.~Morchid, and G.~Linar{\`e}s}, {\em A survey of quaternion
 		neural networks}, Artificial Intelligence Review, 53 (2020), pp.~2957--2982.
 	
 	\bibitem{pei1997novel}
 	{\sc S.-C. Pei and C.-M. Cheng}, {\em A novel block truncation coding of color
 		images using a quaternion-moment-preserving principle}, IEEE Transactions on
 	Communications, 45 (1997), pp.~583--595.
 	
 	\bibitem{pei1999color}
 	{\sc S.-C. Pei and C.-M. Cheng}, {\em Color image processing by using binary
 		quaternion-moment-preserving thresholding technique}, IEEE Transactions on
 	Image Processing, 8 (1999), pp.~614--628.
 	
 	\bibitem{recht2011simpler}
 	{\sc B.~Recht}, {\em A simpler approach to matrix completion.}, Journal of
 	Machine Learning Research, 12 (2011).
 	
 	\bibitem{sangwine1996fourier}
 	{\sc S.~J. Sangwine}, {\em Fourier transforms of colour images using quaternion
 		or hypercomplex, numbers}, Electronics letters, 32 (1996), pp.~1979--1980.
 	
 	\bibitem{shi2007quaternion}
 	{\sc L.~Shi and B.~Funt}, {\em Quaternion color texture segmentation}, Computer
 	Vision and image understanding, 107 (2007), pp.~88--96.
 	
 	\bibitem{song2021low}
 	{\sc G.~Song, W.~Ding, and M.~K. Ng}, {\em Low rank pure quaternion
 		approximation for pure quaternion matrices}, SIAM Journal on Matrix Analysis
 	and Applications, 42 (2021), pp.~58--82.
 	
 	\bibitem{subakan2011quaternion}
 	{\sc {\"O}.~N. Subakan and B.~C. Vemuri}, {\em A quaternion framework for color
 		image smoothing and segmentation}, International Journal of Computer Vision,
 	91 (2011), pp.~233--250.
 	
 	\bibitem{sun2011color}
 	{\sc Y.~Sun, S.~Chen, and B.~Yin}, {\em Color face recognition based on
 		quaternion matrix representation}, Pattern Recognition Letters, 32 (2011),
 	pp.~597--605.
 	
 	\bibitem{tropp2012user}
 	{\sc J.~A. Tropp}, {\em User-friendly tail bounds for sums of random matrices},
 	Foundations of computational mathematics, 12 (2012), pp.~389--434.
 	
 	\bibitem{tropp2015introduction}
 	{\sc J.~A. Tropp}, {\em An introduction to matrix concentration inequalities},
 	arXiv preprint arXiv:1501.01571,  (2015).
 	
 	\bibitem{van2016estimation}
 	{\sc S.~A. van~de Geer}, {\em Estimation and testing under sparsity}, Springer,
 	2016.
 	
 	\bibitem{wainwright2019high}
 	{\sc M.~J. Wainwright}, {\em High-dimensional statistics: A non-asymptotic
 		viewpoint}, vol.~48, Cambridge University Press, 2019.
 	
 	\bibitem{wu2020deep}
 	{\sc J.~Wu, L.~Xu, F.~Wu, Y.~Kong, L.~Senhadji, and H.~Shu}, {\em Deep octonion
 		networks}, Neurocomputing, 397 (2020), pp.~179--191.
 	
 	\bibitem{xu2017multi}
 	{\sc J.~Xu, L.~Zhang, D.~Zhang, and X.~Feng}, {\em Multi-channel weighted
 		nuclear norm minimization for real color image denoising}, in Proceedings of
 	the IEEE international conference on computer vision, 2017, pp.~1096--1104.
 	
 	\bibitem{xu2015vector}
 	{\sc Y.~Xu, L.~Yu, H.~Xu, H.~Zhang, and T.~Nguyen}, {\em Vector sparse
 		representation of color image using quaternion matrix analysis}, IEEE
 	Transactions on image processing, 24 (2015), pp.~1315--1329.
 	
 	\bibitem{yang2021weighted}
 	{\sc L.~Yang, K.~I. Kou, and J.~Miao}, {\em Weighted truncated nuclear norm
 		regularization for low-rank quaternion matrix completion}, arXiv preprint
 	arXiv:2101.02443,  (2021).
 	
 	\bibitem{yang2021low}
 	{\sc L.~Yang, J.~Miao, and K.~I. Kou}, {\em Low rank quaternion matrix recovery
 		via logarithmic approximation}, arXiv preprint arXiv:2107.01380,  (2021).
 	
 	\bibitem{yu2014single}
 	{\sc M.~Yu, Y.~Xu, and P.~Sun}, {\em Single color image super-resolution using
 		quaternion-based sparse representation}, in 2014 IEEE International
 	Conference on Acoustics, Speech and Signal Processing (ICASSP), IEEE, 2014,
 	pp.~5804--5808.
 	
 	\bibitem{yu2019quaternion}
 	{\sc Y.~Yu, Y.~Zhang, and S.~Yuan}, {\em Quaternion-based weighted nuclear norm
 		minimization for color image denoising}, Neurocomputing, 332 (2019),
 	pp.~283--297.
 	
 	\bibitem{zeng2016color}
 	{\sc R.~Zeng, J.~Wu, Z.~Shao, Y.~Chen, B.~Chen, L.~Senhadji, and H.~Shu}, {\em
 		Color image classification via quaternion principal component analysis
 		network}, Neurocomputing, 216 (2016), pp.~416--428.
 	
 	\bibitem{quatpara}
 	{\sc A.~Zhang, Y.~Tay, S.~Zhang, A.~Chan, A.~T. Luu, S.~C. Hui, and J.~Fu},
 	{\em Beyond fully-connected layers with quaternions: Parameterization of
 		hypercomplex multiplications with 1/n parameter}, arXiv preprint
 	arXiv:2102.08597,  (2021).
 	
 	\bibitem{zhang1997quaternions}
 	{\sc F.~Zhang}, {\em Quaternions and matrices of quaternions}, Linear algebra
 	and its applications, 251 (1997), pp.~21--57.
 	
 	\bibitem{quatgraph}
 	{\sc S.~Zhang, Y.~Tay, L.~Yao, and Q.~Liu}, {\em Quaternion knowledge graph
 		embeddings}, arXiv preprint arXiv:1904.10281,  (2019).
 	
 	\bibitem{zhu2018quaternion}
 	{\sc X.~Zhu, Y.~Xu, H.~Xu, and C.~Chen}, {\em Quaternion convolutional neural
 		networks}, in Proceedings of the European Conference on Computer Vision
 	(ECCV), 2018, pp.~631--647.
 	
 \end{thebibliography}

\appendix
\section{Proofs of Technical Tools}
\label{appendixA}
\textbf{Proof of Lemma \ref{lemma1}.} 
Assume $\rank(\bm{A}) = r \leq \min\{M,N\}$, then $\bm{A}$ has r positive singular values $\sigma_1(\bm{A})\geq ...\geq \sigma_r(\bm{A})$. Then we have 
\begin{equation}
    \begin{aligned}
    ||\bm{A}||_{\nuc}= \sum_{i\in[r]} \sigma_i(\bm{A}) \leq \sqrt{(\sum_{i\in[r]} 1^2)(\sum_{i\in[r]}\sigma_i(\bm{A})^2)}\leq \sqrt{r}||\bm{A}||_{\mathrm{F}},
       \nonumber
    \end{aligned}
\end{equation}
where we use Cauchy inequality. \hfill $\square$

\vspace{3mm}
\noindent
\textbf{Proof of Lemma \ref{lemma2}.}
\textbf{I.} We first show that we can only consider square matrices with no loss of generality. If $M<N$, we consider $\widetilde{\bm{A}}= [\bm{A}^T,\bm{0}^T]^T,\ \widetilde{\bm{B}} = [\bm{B}^T,\bm{0}^T]^T\in \mathbb{Q}^{N\times N}$;
If $M>N$ we consider 
$ \widetilde{\bm{A}}= [\bm{A} \ \bm{\mathrm{0}}],\ \widetilde{\bm{B}} = [\bm{B}\  \bm{\mathrm{0}}]\in \mathbb{Q}^{M\times M}$. It can be easily seen that both transforms preserve inner product, nuclear norm and operator norm.

\vspace{2mm}
\noindent
\textbf{II.} Assume $\bm{A,B}\in \mathbb{Q}^{M\times M}$, and $\bm{B} = \bm{U\Sigma V^*}$ is the QSVD of $\bm{B}$, where $\bm{U},\bm{V} \in \mathbb{Q}^{M\times M}$ are unitary, and $\bm{\Sigma} = \mathrm{diag}(\sigma_1,...,\sigma_{M}) \in \mathbb{R}^{M\times M}$. Then 
$\big<\bm{A,B}\big>=\Tr(\bm{A^*B}) = \Tr(\bm{A^*U\Sigma V^*}) = \Tr(\bm{C\Sigma V^*}),$
where we let $\bm{ A^*U}= \bm{C} =[\bm{c_{ij}}]\in \mathbb{Q}^{M\times M}$. Let $\bm{U}= [u_1,...,u_M]$, then the i-th($i\in [M]$) column of $\bm{C}$ is $\bm{A^*}u_i$, so the $\ell_2$ norm of columns of $\bm{C}$ are bounded by $|| \bm{A^*}u_i||_2\leq ||\bm{A^*}||_{\op}||u_i||_2 = ||\bm{A}||_{\op}$. Further let $\bm{V^*}= [\bm{v_{ij}}]$, then we compute element-wisely:
\begin{equation}
\begin{aligned}
&|\Tr(\bm{C\Sigma V^*})| = \Big|\sum_{i\in[M]}\sum_{j\in[M]} \bm{c_{ij}}\sigma_j \bm{v_{ji}}\Big|
= \Big|\sum_j \Big(\sum_i \bm{c_{ij}v_{ji}}\Big) \sigma_j\Big| 
\leq  \sum_j |\sum_i \bm{c_{ij}v_{ji}}| \sigma_j \\
\leq & \sum_j \Big[|| \bm{A^*}u_j||_2^2(\sum_i |\bm{v_{ji}}|^2)\Big]^{1/2}\sigma_j 
\leq  ||\bm{A}||_{\op}\sum_j \sigma_j   =  ||\bm{A}||_{\op} ||\bm{B}||_{\nuc},
\end{aligned}
    \nonumber
\end{equation}
hence the result follows.  \hfill $\square$

\vspace{3mm}
\noindent
\textbf{Proof of Lemma \ref{lemma3}.} We only consider square matrix $\bm{A} \in \mathbb{Q}^{M\times M}$ for the same reason as Lemma \ref{lemma2}. Let $\bm{B} = \bm{A_0 + A_1}\ii,\bm{C} = \bm{A_2} +\bm{A_3}\ii\in \mathbb{C}^{M\times M}$, then $\bm{A} = \bm{B} + \bm{C}\jj$. Denote the set of $M\times M$ quaternion unitary matrices, complex unitary matrices, (real) orthogonal matrices by $\mathcal{U}_{\mathbb{Q}}$, $\mathcal{U}_{\mathbb{C}}$, and $\mathcal{U}_{\mathbb{R}}$ respectively, evidently, $\mathcal{U}_{\mathbb{R}}\subset \mathcal{U}_{\mathbb{C}}\subset\mathcal{U}_{\mathbb{Q}} $. By Theorem 5 in \cite{yang2021low} we have $ ||\bm{A}||_{\nuc} = \sup_{\bm{U,V}\in \mathcal{U}_\mathbb{Q}}|\Tr(\bm{UAV^*}) | ,$
which delivers that 
\begin{equation}
\begin{aligned}
&||\bm{A} ||_{\nuc} \geq  \sup_{\bm{U,V}\in \mathcal{U}_\mathbb{C}}|\Tr(\bm{UAV^*}) | 
=  \sup_{\bm{U,V}\in \mathcal{U}_\mathbb{C}}|\Tr(\bm{U[\bm{B}+ \bm{C}\jj]V^*}) | \\
= & \sup_{\bm{U,V}\in \mathcal{U}_\mathbb{C}}|\Tr(\bm{UBV^*}) + \Tr(\bm{UCV^T})\jj |
\geq  \sup_{\bm{U,V}\in \mathcal{U}_\mathbb{C}} |\Tr(\bm{UBV^*})| =  || \bm{B}||_{\nuc},
\end{aligned} 
    \nonumber
\end{equation}
by complex SVD of $\bm{B}$ it is not hard to show the last equality. We can further proceed it by 
\begin{equation}
    \begin{aligned}
    &|| \bm{B}||_{\nuc} =  \sup_{\bm{U,V}\in \mathcal{U}_\mathbb{C}} |\Tr(\bm{UBV^*})| 
    \geq    \sup_{\bm{U,V}\in \mathcal{U}_\mathbb{R}} |\Tr(\bm{U[\bm{A_0}+ \bm{A_1}\ii]V^*})| \\
    = & \sup_{\bm{U,V}\in \mathcal{U}_\mathbb{R}} | \Tr(\bm{UA_0V^*}) + \Tr(\bm{UA_1V^*})\ii| 
    \geq  \sup_{\bm{U,V}\in \mathcal{U}_\mathbb{R}} | \Tr(\bm{UA_0V^*}) |=||\bm{A_0}||_{\nuc},
    \nonumber
    \end{aligned}
\end{equation}
by real SVD of $\bm{A_0}$ we can verify the last equality. \hfill $\square$

\vspace{3mm}
\noindent
\textbf{Proof of Corollary \ref{corollary1}.} Let $\bm{\nu} $ be the unit quaternion $ \nu_0 - \nu_1\ii - \nu_2\jj -\nu_3\kk$, we consider $\bm{\nu \bm{A}}$, then we have $||\bm{\nu \bm{A}}||_{\nuc} = ||\bm{A} ||_{\nuc} $. Evidently, we have $[\bm{\nu \bm{A}} ]_{\rr} = \nu_0 \bm{A_0 } + \nu_1\bm{A_1}+ \nu_2\bm{A_2}+ \nu_3 \bm{A_3}, $
then the result follows by using Lemma \ref{lemma3}. \hfill $\square$

\vspace{3mm}
\noindent
\textbf{Proof of Lemma \ref{lemma4}.} The main ingredient is the complex adjoint matrix (\ref{2.2}). We consider the complex adjoint $\{[\bm{S_1}]_{\mathbb{C}},...,[\bm{S_n}]_{\mathbb{C}}\}\subset \mathbb{C}^{2M\times 2M}$, which are independent Hermitian random matrices with zero mean. Note that $[.]_{\mathbb{C}}$ preserves the operator norm, expectation, and multiplication, so for any integer $p\geq 2$ we have 
\begin{equation}
    \begin{aligned}
    \frac{1}{2}p!R^{p-2}\sigma^2 \geq  ||\mathbbm{E} \bm{S_k^p} ||_{\op}  =  || [\mathbbm{E}\bm{S_k^p}]_{\mathbb{C}}||_{\op} = || \mathbbm{E} [\bm{S_k^p}]_{\mathbb{C}}||_{\op} = || \mathbbm{E} [\bm{S_k}]_{\mathbb{C}}^p||_{\op} . 
    \nonumber
    \end{aligned}
\end{equation}
Then by using Theorem 6.2 in \cite{tropp2012user}, for any $t>0$ we have 
\begin{equation}
    \begin{aligned}
    &\mathbbm{P}\Big[||\frac{1}{n}\sum_{k=1}^n\bm{S_k} ||_{\op} \geq t\Big] =  \mathbbm{P}\Big[||\sum_{k=1}^n [\bm{S_k}]_{\mathbb{C}} ||_{\op} \geq nt\Big] \\ 
    \leq & \sum_{a=\pm 1}\mathbbm{P}\Big[\lambda_{\max}(\sum_{k=1}^n [a\bm{S_k}]_{\mathbb{C}}) \geq nt\Big]  \leq  4M\exp\left(-\frac{nt^2}{2(\sigma^2+ Rt)}\right),
    \nonumber
    \end{aligned}
\end{equation}
which finishes the proof. \hfill $\square$
\section{Proofs of Main Results}
\label{appendixB}
\subsection{Error Bound}
\textbf{Proof of Lemma \ref{lemma5}.} \textbf{I.} We first decompose the complementary event into countably infinite events. State the complementary event as 
$$\mathscr{B}= \{\exists\  \bm{\Theta_0}\in \mathcal{C}(\psi),\ \mathrm{s.t.}\ \mathcal{F}_{\mathscr{X}}(\bm{\Theta_0})\leq \frac{\kappa}{d_1d_2}|| \bm{\Theta_0}||_{\mathrm{w},\mathrm{F}}^2- T_0\} ,$$
note that $\mathbbm{E} \mathcal{F}_{\mathscr{X}}(\bm{\Theta})= || \bm{\Theta}||_{\mathrm{w},\mathrm{F}}^2/(d_1d_2)$, so $\mathscr{B}$ implies 
\begin{equation}
|\mathcal{F}_{\mathscr{X}}(\bm{\Theta_0})-\mathbbm{E}\mathcal{F}_{\mathscr{X}}(\bm{\Theta_0})| \geq \left(\frac{1-\kappa}{d_1d_2}\right)||\bm{\Theta_0}||^2_{\mathrm{w,F}}+ T_0.
    \label{B.1}
\end{equation}
By (\ref{3.9}) we know $||\bm{\Theta_0}||^2_{\mathrm{w,F}}\geq D_0 =: \alpha^2 d_1d_2\sqrt{\frac{\psi_3\log(d_1+d_2)}{n}},$
hence by a fixed $\zeta >1$ (which will be specified later) there exists positive integer $l$ such that $ || \bm{\Theta_0}||_{\mathrm{w,F}}^2 \in [\zeta^{l-1}D_0, \zeta^lD_0)$, hence \begin{equation}
    \bm{\Theta_0}\in \mathcal{C}(\psi,l) := \mathcal{C}(\psi)\cap \{\bm{\Theta}:  || \bm{\Theta}||^2_{\mathrm{w,F}}\in [ \zeta^{l-1}D_0, \zeta^lD_0)  \}.
    \label{B.2}
\end{equation}
We define $\mathcal{Z}_{\mathscr{X}}(l) = \sup_{\bm{\Theta}\in \mathcal{C}(\psi, l)} |\mathcal{F}_{\mathscr{X}}(\bm{\Theta})-\mathbbm{E}\mathcal{F}_{\mathscr{X}}(\bm{\Theta})|, $
then by combining (\ref{B.1}), (\ref{B.2}) we know the event defined by \begin{equation}
\mathscr{B}_l := \Big\{ \mathcal{Z}_{\mathscr{X}}(l )  \geq \left(\frac{1-\kappa}{d_1d_2}\right)\zeta^{l-1}D_0+ T_0\Big\}
    \label{B.3}
\end{equation}
holds. Therefore, we have 
\begin{equation}
\mathbbm{P}(\mathscr{B}) \leq \sum_{l=1}^{\infty} \mathbbm{P}(\mathscr{B}_l).
    \label{B.4}
\end{equation}

\vspace{2mm}
\noindent
\textbf{II.} It suffices to bound $\mathbbm{P}(\mathscr{B}_l)$ for positive integer $l$. We first bound the deviation $|\mathcal{Z}_{\mathscr{X}}(l)-\mathbbm{E}\mathcal{Z}_{\mathscr{X}}(l)|$. Let $\widetilde{\mathscr{X}} = \{ \bm{\widetilde{X}_1}, \bm{X_2,...,\bm{X_n}}\}$, i.e., only the first component may vary from $\mathscr{X}$, then for any positive integer $l $ we have 
\begin{equation}
    \begin{aligned}
    &\sup_{\mathscr{X},\widetilde{\mathscr{X}}}|\mathcal{Z}_{\mathscr{X}}(l)-\mathcal{Z}_{\widetilde{\mathscr{X}}}(l)|  
    =  \sup_{\mathscr{X},\widetilde{\mathscr{X}}} \Big|\sup_{\bm{\Theta}\in \mathcal{C}(\psi, l)} |\mathcal{F}_{\mathscr{X}}(\bm{\Theta})-\mathbbm{E}\mathcal{F}_{\mathscr{X}}(\bm{\Theta})|-\sup_{\bm{\Theta}\in \mathcal{C}(\psi, l)}  |\mathcal{F}_{\widetilde{\mathscr{X}}}(\bm{\Theta})-\mathbbm{E}\mathcal{F}_{\widetilde{\mathscr{X}}}(\bm{\Theta})| \Big| \\ &\leq  \sup_{\mathscr{X},\widetilde{\mathscr{X}}}\Big|\sup_{\bm{\Theta}\in \mathcal{C}(\psi, l)} |\mathcal{F}_{\mathscr{X}}(\bm{\Theta})-\mathcal{F}_{\widetilde{\mathscr{X}}}(\bm{\Theta}) | \Big| 
    =  \sup_{\bm{X_1},\bm{\widetilde{X}_1}} \sup_{\bm{\Theta}\in \mathcal{C}(\psi, l)} \frac{1}{n}\Big||\big< \bm{X_1},\bm{\Theta}\big>|^2_{\mathrm{w}} -|\big< \bm{\widetilde{X}_1},\bm{\Theta}\big>|^2_{\mathrm{w}}\Big| 
    \leq \frac{\psi_1^2\alpha^2}{n},
    \nonumber
    \end{aligned}
\end{equation}
where we use $||\bm{\Theta}||_{\mathrm{w},\infty} \leq \psi_1\alpha$ in the last inequality. Since n components in $\mathscr{X}$ are symmetrical, by bounded difference inequality (e.g., Corollary 2.21, \cite{wainwright2019high}), for any $t>0$ we have 
\begin{equation}
\mathbbm{P}\Big[\mathcal{Z}_{\mathscr{X}}(l)- \mathbbm{E}\mathcal{Z}_{\mathscr{X}}(l) \geq t \Big] \leq \exp \left(- \frac{2nt^2}{\psi_1^4 \alpha^4} \right).
    \label{B.5}
\end{equation}
It remains to bound $\mathbbm{E}\mathcal{Z}_{\mathscr{X}}(l)$. Let $\mathscr{E}=(\varepsilon_1,...,\varepsilon_n)$ be i.i.d. Rademacher random variables ($\mathbbm{P}[\varepsilon_k = 1] = \mathbbm{P}[\varepsilon_k = -1]= 1/2 $), then by symmetrization of expectations (e.g., Theorem 16.1, \cite{van2016estimation}), we plug in the definition of $\mathcal{Z}_{\mathscr{X}}(l)$, $\mathcal{F}_{\mathscr{X}}(\bm{\Theta})$, it yields that
\begin{equation}
\begin{aligned}
&\mathbbm{E}\mathcal{Z}_{\mathscr{X}}(l)  =\mathbbm{E} \sup_{\bm{\Theta}\in \mathcal{C}(\psi, l)} \Big|\frac{1}{n}\sum_{k=1}^n \Big\{|\big<\bm{X_k,\Theta}\big>|_{\mathrm{w}}^2 -\mathbbm{E}|\big<\bm{X_k,\Theta}\big>|_{\mathrm{w}}^2\Big\}  \Big| \\ 
\leq &  2\mathbbm{E} \sup_{\bm{\Theta}\in \mathcal{C}(\psi, l)} \Big| \frac{1}{n}\sum_{k=1}^n \varepsilon_k |\big< \bm{X_k,\Theta}\big>|_{\mathrm{w}}^2\Big| 
=  2\mathbbm{E}_{\mathscr{X}}\mathbbm{E}_{\mathscr{E}} \sup_{\bm{\Theta}\in \mathcal{C}(\psi, l)} \Big| \frac{1}{n}\sum_{k=1}^n \varepsilon_k |\big< \bm{X_k,\sqrt{\bm{W}}\Theta}\big>|^2\Big|,
\label{B.6}
\end{aligned}
\end{equation}
where $\mathbbm{E}_{\mathscr{X}}$, $\mathbbm{E}_{\mathscr{E}}$ denote taking expectation of the random variables in subscript. Moreover, by dividing three parts, applying Talagrand's inequality (e.g., Theorem 16.2, \cite{van2016estimation}) we have
\begin{equation}
\begin{aligned}
&\mathbbm{E}_{\mathscr{E}}\sup_{\bm{\Theta}\in \mathcal{C}(\psi, l)} \Big| \frac{1}{n}\sum_{k=1}^n \varepsilon_k |\big< \bm{X_k,\sqrt{\bm{W}}\Theta}\big>|^2\Big| \\
\leq & (2\psi_1\alpha)^2\sum_{\vartheta \in \{\ii,\jj,\kk\}} \mathbbm{E}_{\mathscr{E}} \sup_{\bm{\Theta}\in \mathcal{C}(\psi, l)} \Big|\frac{1}{n}\sum_{k=1}^n \varepsilon_k \big< \bm{X_k}, [\bm{\sqrt{W}\Theta}]_{\vartheta}(2\psi_1\alpha)^{-1}\big>^2 \Big| \\ \leq & 4\psi_1\alpha \sum_{\vartheta \in \{\ii,\jj,\kk\}} \mathbbm{E}_{\mathscr{E}} \sup_{\bm{\Theta}\in \mathcal{C}(\psi, l)} \Big| \big<\frac{1}{n}\sum_{k=1}^n \varepsilon_k \bm{X_k} , [\bm{\sqrt{W}\Theta}]_{\vartheta} \big>\Big|.
\end{aligned}
    \label{B.7}
\end{equation}
By plugging (\ref{B.7}) in (\ref{B.6}) and using Corollary \ref{corollary1}, it yields that 
\begin{equation}
\begin{aligned}
\mathbbm{E} \mathcal{Z}_{\mathscr{X}}& (l)\leq 8\psi_1 \alpha \sum_{\vartheta \in \{\ii,\jj,\kk\}} \Big\{\mathbbm{E} \Big|\Big|\frac{1}{n}\sum_{k=1}^n\varepsilon_k \bm{X_k}\Big|\Big|_{\op} \Big\}\sup_{\bm{\Theta}\in \mathcal{C}(\psi, l)} \Big|\Big| [\bm{\sqrt{W}\Theta}]_{\vartheta}\Big|\Big|_{\nuc}\\
\leq & 24\psi_1 \alpha \Big\{ \mathbbm{E} \Big|\Big|\frac{1}{n}\sum_{k=1}^n\varepsilon_k \bm{X_k}\Big|\Big|_{\op}\Big\}||\sqrt{\bm{W}} ||_{2,\infty}\sup_{\bm{\Theta}\in \mathcal{C}(\psi, l)}|| \bm{\Theta}||_{\nuc}
\\\leq & 24\psi_1 \psi_2 \rho^{\frac{1}{2-q}}\alpha\Big\{  \mathbbm{E} \Big|\Big|\frac{1}{n}\sum_{k=1}^n\varepsilon_k \bm{X_k}\Big|\Big|_{\op}\Big\}||\sqrt{\bm{W}} ||_{2,\infty}\sup_{\bm{\Theta}\in \mathcal{C}(\psi, l)}|| \bm{\Theta} ||_{\mathrm{F}}^{\frac{2-2q}{2-q}} ,
\end{aligned}
    \label{B.8}
\end{equation}
while clearly we have $||\bm{\sqrt{W}} ||_{2,\infty}= \sqrt{||\bm{W}||_{\infty}}$. By Matrix Bernstein (Theorem 6.1.1, \cite{tropp2015introduction}) it is not hard to show (assume $[\min\{d_1,d_2\}\log(d_1+d_2)]/n\leq 1/16$)
\begin{equation}
\mathbbm{E} \Big|\Big|\frac{1}{n}\sum_{k=1}^n\varepsilon_k \bm{X_k}\Big|\Big|_{\op} \leq \frac{3}{2}(d_1d_2)^{-\frac{1}{2}}\sqrt{\frac{\max\{d_1,d_2\}\log(d_1+d_2)}{n}}.
    \label{B.9}
\end{equation}
By (\ref{B.2}) we know for any $\bm{\Theta}\in \mathcal{C}(\psi, l)$ it holds that 
\begin{equation}
\zeta^lD_0 \geq || \bm{\Theta}||_{\mathrm{w,F}}^2 \geq \lambda_{\min}(\bm{W}) || \bm{\Theta}||_{\mathrm{F}}^2
    \label{B.10}
\end{equation}
Plug (\ref{B.9}), (\ref{B.10}) in (\ref{B.8}), some algebra yields 
\begin{equation}
\begin{aligned}
\mathbbm{E} \mathcal{Z}_{\mathscr{X}} (l) \leq \Big\{ (2-q)T_0\Big\}^{\frac{1}{2-q}}\Big\{\frac{\zeta^lD_0}{d_1d_2}\Big\}^{\frac{1-q}{2-q}} \leq \frac{1-q}{2-q}\frac{\zeta^lD_0}{d_1d_2}+ T_0
\end{aligned}
    \label{B.11}
\end{equation}
By combining with (\ref{B.3}), (\ref{B.5}), we have 
\begin{equation}
    \begin{aligned}
    \mathbbm{P}(\mathscr{B}_l) \leq & \mathbbm{P}\Big[\mathcal{Z}_{\mathscr{X}}(l) -\mathbbm{E}\mathcal{Z}_{\mathscr{X}}(l) \geq  \left(\frac{1-\kappa}{\zeta}- \frac{1-q}{2-q}\right)\frac{\zeta^l D_0}{d_1d_2}\Big] 
    \leq  \exp\left(-\frac{2n\kappa_1^2\zeta^{2l}D_0^2}{\psi_1^4\alpha^4(d_1d_2)^2}\right),
    \end{aligned}
\end{equation}
where we let $\kappa_1 = \frac{1-\kappa}{\zeta} - \frac{1-q}{2-q}$, which is in $(0,1)$ by letting $\kappa$ close to 0, $\zeta$ close to 1. We then take the union bound, plug in $D_0 $, and use $\zeta^{2l}> 2l \log \zeta$
\begin{equation}
\begin{aligned}
 \mathbbm{P}(\mathscr{B}) \leq  \sum_{l=1}^{\infty} \mathbbm{P}(\mathscr{B}_l) 
 \leq \sum_{l=1}^{\infty}(d_1+d_2)^{-\frac{4\psi_3 \kappa_1^2  \log \zeta }{\psi_1^4} l} 
 \leq 2(d_1+d_2)^{-3/4},
\end{aligned}
    \label{B.13}
\end{equation}
the last inequality holds if we let $\psi_3 \geq  \frac{3\psi_1^4}{16\kappa_1^2 \log \zeta}.$ \hfill $\square$

\vspace{3mm}
\noindent
\textbf{Proof of Lemma \ref{lemma6}.} \textbf{(1) I.} We first introduce a pair of subspaces $(\mathcal{M},\overline{\mathcal{M}})$ and build some useful inequalities. Write the QSVD of $\bm{\widetilde{\Theta}}$ as 
\begin{equation}
\bm{U\Sigma V^*} = [\bm{U_1}\ \bm{U_2}] \begin{bmatrix} \bm{\Sigma_{11}}  & \bm{0}\\ \bm{0}& \bm{\Sigma_{22}}\end{bmatrix} \begin{bmatrix} \bm{V_1^*} \\ \bm{V_2^*} \end{bmatrix},
    \label{B.14}
\end{equation}
where $\bm{U_1}\in \mathbb{Q}^{d_1 \times z}$, $\bm{U_2}\in \mathbb{Q}^{d_1 \times (d_1-z)}$ are partition of left singular vectors, $\bm{V_1}\in \mathbb{Q}^{d_2\times z}$, $\bm{V_2}\in \mathbb{Q}^{d_2 \times (d_2 -z)}$ are partition of right singular vectors, the diagonal of $\bm{\Sigma_{11}} \in \mathbb{R}^{z\times z}$ contains the z largest singular values, the remaining $\min\{d_1,d_2\}-z$ singular values are arranged in $\bm{\Sigma_{22}}$, here the integer $z\in \{0,1,...,\min\{d_1,d_2\}\}$ will be selected later. Corresponding to a specific $z$ we define the subspace $\mathcal{M},\overline{\mathcal{M}}\in \mathbb{Q}^{d_1\times d_2}$ by 
$\mathcal{M} = \{ \bm{U_1A_1V_1^*}: \bm{A_1}\in \mathbb{Q}^{z\times z} \}$ and
$$\overline{\mathcal{M}}= \Big\{  [\bm{U_1}\ \bm{U_2}] \begin{bmatrix} \bm{A_1}  & \bm{A_2} \\ \bm{A_3}&\bm{0}\end{bmatrix} \begin{bmatrix} \bm{V_1^*} \\ \bm{V_2^*} \end{bmatrix}: \bm{A_1}\in \mathbb{Q}^{z\times z},\bm{A_2}\in \mathbb{Q}^{z\times (d_1-z)},\bm{A_3}\in \mathbb{Q}^{(d_1 -z)\times z}\Big\}.$$
We use $\mathcal{P}_{\mathcal{M}}$, $\mathcal{P}_{\overline{\mathcal{M}}}$ to denote the projection onto $\mathcal{M}$, $\overline{\mathcal{M}}$ respectively. Consider any $\bm{\Delta}\in \mathbb{Q}^{d_1\times d_2}$
$$\bm{\Delta} = [\bm{U_1}\ \bm{U_2}] \begin{bmatrix} \bm{\Delta_{11}}  &\bm{\Delta_{12}} \\ \bm{\Delta_{21}}&\bm{\Delta_{22}}\end{bmatrix} \begin{bmatrix} \bm{V_1^*} \\ \bm{V_2^*} \end{bmatrix}, $$
$\mathcal{P}_{\mathcal{M}}$ and $\mathcal{P}_{\overline{\mathcal{M}}}$ can be given explicitly as 
$$\mathcal{P}_{\mathcal{M}}\bm{\Delta}= \bm{U_1\Delta_{11}\bm{V_1^*}}\ ; \ \mathcal{P}_{\overline{\mathcal{M}}}\bm{\Delta} = [\bm{U_1}\ \bm{U_2}] \begin{bmatrix} \bm{\Delta_{11}}  &\bm{\Delta_{12}} \\ \bm{\Delta_{21}}& \bm{0}\end{bmatrix} \begin{bmatrix} \bm{V_1^*} \\ \bm{V_2^*} \end{bmatrix}.$$
Besides, we let $\mathcal{P}_{\mathcal{M}^\bot}\bm{\Delta} = \bm{\Delta}-\mathcal{P}_{\mathcal{M}}\bm{\Delta} $, $\mathcal{P}_{\overline{\mathcal{M}}^{\bot}} \bm{\Delta}= \bm{\Delta} - \mathcal{P}_{\overline{\mathcal{M}}} \bm{\Delta} $. Note that nuclear norm is decomposable \cite{negahban2012restricted} with respect to $(\mathcal{M}, \overline{\mathcal{M}})$ in the sense that 
\begin{equation}
 || \mathcal{P}_{\mathcal{M}}\bm{\Delta_1}+ \mathcal{P}_{\overline{\mathcal{M}}^\bot}\bm{\Delta_2}||_{\nuc} = ||\mathcal{P}_{\mathcal{M}}\bm{\Delta_1} ||_{\nuc } + || \mathcal{P}_{\overline{\mathcal{M}}^\bot}\bm{\Delta_2}||_{\nuc } ,\ \forall \bm{\Delta_1,\Delta_2}\in \mathbb{Q}^{d_1\times  d_2}.
    \label{B.15}
\end{equation}
Let $\sigma_1(\bm{\widetilde{\Theta}}) \geq ...\geq \sigma_{\min\{d_1,d_2\}}(\bm{\widetilde{\Theta}})$ be the singular values of $\bm{\widetilde{\Theta}}$, we specify a threshold $\tau >0$, and then determine $z$ by $z = \max \Big\{ \{0\}\cup \{i\in [\min\{d_1,d_2\}]:\sigma_i(\bm{\widetilde{\Theta}}) \geq \tau \}  \Big\}.$
Recall the approximate low-rankness (\ref{3.8}), we have 
\begin{equation}
    \rho \geq \sum_{i=1}^z \sigma_i(\bm{\widetilde{\Theta}})^q \geq \tau^qz \Longrightarrow z\leq \rho \tau^{-q}.
    \label{B.16}
\end{equation}
For the $\min\{d_1,d_2\}-z$ smallest singular values we have 
\begin{equation}
||\mathcal{P}_{\mathcal{M}^\bot} \bm{\widetilde{\Theta}} ||_{\nuc} = \sum_{i=z+1}^{\min\{d_1,d_2\}} \sigma_i(\bm{\widetilde{\Theta}})^{q}\sigma_i(\bm{\widetilde{\Theta}})^{1-q} \leq \rho \tau^{1-q}.
    \label{B.17}
\end{equation}
\textbf{II.} We use the established $(\mathcal{M},\overline{\mathcal{M}})$ to proceed. By decomposability (\ref{B.15}) and triangle inequlity, we have 
\begin{equation}
   \begin{aligned}
     &||\bm{\widehat{\Theta}}||_{\mathrm{nu}} - ||\bm{\widetilde{\Theta}}||_{\mathrm{nu}} 
        = ||\mathcal{P}_{\mathcal{M}}\bm{\widetilde{\Theta}}+\mathcal{P}_{\mathcal{M}^\perp}\bm{\widetilde{\Theta}} +\mathcal{P}_{\overline{\mathcal{M}}}\bm{\widehat{\Delta}}+\mathcal{P}_{\overline{\mathcal{M}}^\perp}\bm{\widehat{\Delta}}||_{\mathrm{nu}}-||\mathcal{P}_{\mathcal{M}}\bm{\widetilde{\Theta}}+\mathcal{P}_{\mathcal{M}^\perp}\bm{\widetilde{\Theta}} ||_{\mathrm{nu}} \\
        \geq & ||\mathcal{P}_{\mathcal{M}}\bm{\widetilde{\Theta}}||_{\mathrm{nu}} +||\mathcal{P}_{\overline{\mathcal{M}}^\perp}\bm{\widehat{\Delta}}||_{\mathrm{nu}}-||\mathcal{P}_{\mathcal{M}^\perp}\bm{\widetilde{\Theta}}||_{\mathrm{nu}}-||\mathcal{P}_{\overline{\mathcal{M}}}\bm{\widehat{\Delta}}||_{\mathrm{nu}} -||\mathcal{P}_{\mathcal{M}}\bm{\widetilde{\Theta}}||_{\mathrm{nu}} -||\mathcal{P}_{\mathcal{M}^\perp}\bm{\widetilde{\Theta}}||_{\mathrm{nu}}\\
        = & ||\mathcal{P}_{\overline{\mathcal{M}}^\perp}\bm{\widehat{\Delta}}||_{\mathrm{nu}}-2||\mathcal{P}_{\mathcal{M}^\perp}\bm{\widetilde{\Theta}}||_{\mathrm{nu}}-||\mathcal{P}_{\overline{\mathcal{M}}}\bm{\widehat{\Delta}}||_{\mathrm{nu}}.
   \end{aligned}
        \label{B.18}
\end{equation}
By combining with the optimality in (\ref{3.4}) we have $||\mathcal{P}_{\overline{\mathcal{M}}^\perp}\bm{\widehat{\Delta}}||_{\mathrm{nu}} \leq 2||\mathcal{P}_{\mathcal{M}^\perp}\bm{\widetilde{\Theta}}||_{\mathrm{nu}}+||\mathcal{P}_{\overline{\mathcal{M}}}\bm{\widehat{\Delta}}||_{\mathrm{nu}}$ then we bound $||\bm{\widehat{\Delta}} ||_{\nuc}$ by triangle inequality, (\ref{B.17}), Lemma \ref{lemma1}, $\rank(\mathcal{P}_{\overline{\mathcal{M}}}\bm{\widehat{\Delta}})\leq 2z$, (\ref{B.16})
\begin{equation}
    \begin{aligned}
    ||\bm{\widehat{\Delta}} ||_{\nuc} \leq & ||\mathcal{P}_{\overline{\mathcal{M}}}\bm{\widehat{\Delta}} ||_{\nuc}   +||\mathcal{P}_{\overline{\mathcal{M}}^\bot}\bm{\widehat{\Delta}} ||_{\nuc}  \leq  2||\mathcal{P}_{\mathcal{M}^\perp}\bm{\widetilde{\Theta}}||_{\mathrm{nu}}+2||\mathcal{P}_{\overline{\mathcal{M}}}\bm{\widehat{\Delta}}||_{\mathrm{nu}} 
   \\ \leq & 2[\rho\tau^{1-q} + \sqrt{2\rho}\tau^{-q/2}||\bm{\widehat{\Delta}} ||_{\mathrm{F}}]
      \label{B.19}
    \end{aligned}
\end{equation}
which holds for any threshold $\tau>0$. Assume $\bm{\widehat{\Delta}}\neq 0$, we choose $ \tau = \left(\frac{||\bm{\widehat{\Delta}} ||_{\mathrm{F}}^2}{\rho}\right)^{\frac{1}{2-q}}$
and then obtain (\ref{3.13}).

\vspace{2mm}
\noindent
\textbf{(2)} Again we use the framework established in the proof of (1). From the optimality of (\ref{3.7}), we have 
\begin{equation}
\lambda(|| \bm{\widehat{\Theta}}||_{\nuc} - || \bm{\widetilde{\Theta}}||_{\nuc} ) \leq \mathcal{L}_{\mathrm{w}}(\bm{\widetilde{\Theta}}) - \mathcal{L}_{\mathrm{w}}(\bm{\widehat{\Theta}}).
    \label{B.20}
\end{equation}
Recall the notation in (\ref{3.10}), some algebra yields 
\begin{equation}
\mathcal{L}_{\mathrm{w}}(\bm{\widehat{\Theta}}) - \mathcal{L}_{\mathrm{w}}(\bm{\widetilde{\Theta}}) = \frac{1}{2} \mathcal{F}_{\mathscr{X}}(\bm{\widehat{\Delta}} ) - \mathrm{Re}\big<\frac{1}{n}\sum_{k=1}^n (\bm{W\epsilon_k})\bm{X_k}, \bm{\widehat{\Delta}}\big>.
    \label{B.21}
\end{equation}
Further use Lemma \ref{lemma2} and our choice of $\lambda$ (\ref{3.14}), we obtain
\begin{equation}
    \begin{aligned}
    \mathcal{L}_{\mathrm{w}}(\bm{\widetilde{\Theta}}) - \mathcal{L}_{\mathrm{w}}(\bm{\widehat{\Theta}}) \leq \Big| \Big| \frac{1}{n}\sum_{k=1}^n (\bm{W\epsilon_k})\bm{X_k}\Big| \Big|_{\op} || \bm{\widehat{\Delta}}||_{\nuc} \leq \frac{\lambda}{C_1}||\bm{\widehat{\Delta}} ||_{\nuc}.
    \label{B.22}
    \end{aligned}
\end{equation}
Plug $||\bm{\widehat{\Delta}} ||_{\nuc}\leq ||\mathcal{P}_{\overline{\mathcal{M}}} \bm{\widehat{\Delta}} ||_{\nuc} + ||\mathcal{P}_{\overline{\mathcal{M}}^\bot} \bm{\widehat{\Delta}} ||_{\nuc} $ in (\ref{B.22}), then combine with (\ref{B.20}),
and further use (\ref{B.18}) which is still valid here, we obtain 
$$||\mathcal{P}_{\overline{\mathcal{M}}^\bot} \bm{\widehat{\Delta}} ||_{\nuc} \leq \frac{C_1+1}{C_1- 1} ||\mathcal{P}_{\overline{\mathcal{M}}} \bm{\widehat{\Delta}} ||_{\nuc}  + \frac{2C_1}{C_1-1}||\mathcal{P}_{\mathcal{M}^\bot}\bm{\widetilde{\Theta}} ||_{\nuc},$$
again we use triangle inequality, it yields that 
\begin{equation}
||\bm{\widehat{\Delta}} ||_{\nuc}\leq ||\mathcal{P}_{\overline{\mathcal{M}}} \bm{\widehat{\Delta}} ||_{\nuc} + ||\mathcal{P}_{\overline{\mathcal{M}}^\bot} \bm{\widehat{\Delta}} ||_{\nuc} \leq \frac{2C_1}{C_1-1}\Big[||\mathcal{P}_{\overline{\mathcal{M}}} \bm{\widehat{\Delta}} ||_{\nuc}  + ||\mathcal{P}_{\mathcal{M}^\bot}\bm{\widetilde{\Theta}} ||_{\nuc} \Big]
    \label{B.23}
\end{equation}
Then we follow (\ref{B.19}) and the choice of $\tau$ in (1), (\ref{3.15}) follows. \hfill $\square$

\vspace{3mm}
\noindent
\textbf{Proof of Theorem \ref{theorem1}.} By (\ref{3.3}), (\ref{3.4}) we know $|| \bm{\widehat{\Delta}}||_{\infty}\leq 2\alpha$. Also, recall that $\bm{\widehat{\Delta}}$ satisfies (\ref{3.13}). Let $\bm{W=I_3}$. We consider the constraint set $\mathcal{C}(2,5,\psi_3)$ where $\psi_3$ are slightly large such that Lemma \ref{lemma5} holds, i.e., there exists $\kappa \in (0,1)$ such that with probability higher than $1-2(d_1+d_2)^{-3/4}$, 
\begin{equation}
\frac{1}{n}\sum_{k=1}^n |\big< \bm{X_k,\Theta}\big> |^2 \geq \kappa \frac{|| \bm{\Theta}||_{\mathrm{F}}^2}{d_1d_2} - T_0
    \label{B.24}
\end{equation}
holds for all $\bm{\Theta}\in \mathcal{C}(2,5,\psi_3)$. By constraint in (\ref{3.4}) we know 
$\frac{1}{n}\sum_{k=1}^n |\big<\bm{X_k,\widehat{\Delta} } \big>|^2=0.$
We rule out $2(d_1+d_2)^{-3/4}$ probability and assume (\ref{B.24}) holds. Let us obtain (\ref{3.16}) by considering two possible cases.

\vspace{2mm}
\noindent
\textbf{Case 1.} $\bm{\widehat{\Delta}}\in \mathcal{C}(2,5,\psi_3)$. By (\ref{B.24}) we obtain $\kappa || \bm{\widehat{\Delta}}||^2_{\mathrm{F}}/(d_1d_2) \leq T_0.$
Note that we let $\bm{W}= \bm{I_3}$, by plugging in (\ref{3.12}) we obtain (\ref{3.16}).

\vspace{2mm}
\noindent
\textbf{Case 2.} $\bm{\widehat{\Delta}}\notin \mathcal{C}(2,5,\psi_3)$. Since we have showed that $\bm{\widehat{\Delta}}$ satisfies the first, the second constraint of $\mathcal{C}(2,5,\psi_3)$, hence it violates the third one, i.e.,
$\frac{||\bm{\widehat{\Delta}}||_{\mathrm{F}}^2}{d_1d_2}\leq \alpha^2\sqrt{\frac{\psi_3\log(d_1+d_2)}{n}},$
which also implies (\ref{3.16}) under the scaling $n<d_1d_2$ and $\rho \gtrsim (d_1d_2)^{q/2}$.\hfill $\square$

\vspace{3mm}
\noindent
\textbf{Proof of Lemma \ref{lemma7}.} We let $\bm{\tilde{\epsilon}_k}=\bm{W}\bm{\epsilon_k}$, then $\bm{\tilde{\epsilon}_k}\sim \mathcal{N}(0,\bm{W\Sigma_{c} W})$ if viewed as 3-dimensional real random vector. To simplify the notation, we define $\bm{\widetilde{\Sigma}_c}$ to be $\bm{W\Sigma_c W}$. We aim to use Lemma \ref{lemma4} to show the concentration property of $||\sum_{k=1}^n \bm{\tilde{\epsilon}_kX_k}/n ||_{\op}$, hence we first transform the rectangular matrix $\bm{\tilde{\epsilon}_kX_k}$ to square matrix:
$$\bm{S_k} = \begin{bmatrix} \bm{0} & \bm{\tilde{\epsilon}_kX_k} \\
\bm{\tilde{\epsilon}_k^*X_k^T} & \bm{0}\end{bmatrix}\in \mathbb{Q}^{(d_1+d_2)\times (d_1+d_2)}.$$
Note that $\bm{S_1,...,S_n}$ are independent Hermitian random matrices with zero mean, i.e., $\mathbbm{E}\bm{S_k} = \bm{0}$. Besides, it is not hard to show
$\sum_{k=1}^n\bm{S_k} $ has the same operator norm as $\sum_{k=1}^n \bm{\tilde{\epsilon}_kX_k}$, then it remains to bound the moments of $\mathbbm{E}\bm{S_k^l}$ for $l\geq2$. By calculation, for any $p\geq 1$, we have 
$$\bm{S_k}^{2p}=  \begin{bmatrix} \bm{|\tilde{\epsilon}_k|^{2p}X_kX_k^T} & 0 \\ 0 &   \bm{|\tilde{\epsilon}_k|^{2p}X_k^TX_k} \end{bmatrix}.$$
We know $\bm{X_k}= e_{k(i)}e_{k(j)}^T$ and $(k(i),k(j)) \sim \mathrm{unif}([d_1]\times [d_2])$, hence $\bm{X_kX_k^T} = e_{k(i)}e_{k(i)}^T \in \mathbb{R}^{d_1\times d_1}$, which implies that $\mathbbm{E}\bm{X_kX_k^T}= d_1^{-1}\bm{I_{d_1}}$. Similarly, we have $\mathbbm{E}\bm{X_k^TX_k}= d_2^{-1}\bm{I_{d_2}}$. Thus, $||\mathbbm{E}\bm{S_k^{2p}}||_{\op} =  \mathbbm{E}\bm{|\tilde{\epsilon}_k|^{2p}}/\min\{d_1,d_2\}.$
Since $\bm{\tilde{\epsilon}_k} \sim \mathcal{N}(\bm{0}, \bm{\widetilde{\Sigma}_{c}})$, assume the diagonal of $\bm{\widetilde{\Sigma}_c}$ is $(\tilde{\sigma}_{11}^2,\tilde{\sigma}_{22}^2,\tilde{\sigma}_{33}^2)^T$, then we know $[\bm{\tilde{\epsilon}_k}]_{\ii}\sim \mathcal{N}(0,\tilde{\sigma}_{11}^2)$, $[\bm{\tilde{\epsilon}_k}]_{\jj}\sim \mathcal{N}(0,\tilde{\sigma}_{22}^2)$, $[\bm{\tilde{\epsilon}_k}]_{\kk}\sim \mathcal{N}(0,\tilde{\sigma}_{33}^2)$ 
\begin{equation}
    \begin{aligned}
    \mathbbm{E}\bm{|\tilde{\epsilon}_k|^{2p}} = & \mathbbm{E} [[\bm{\tilde{\epsilon}_k}]_{\ii}^2 + [\bm{\tilde{\epsilon}_k}]_{\jj}^2+[\bm{\tilde{\epsilon}_k}]_{\kk}^2]^p 
    \leq  3^{p-1}\mathbbm{E}[[\bm{\tilde{\epsilon}_k}]_{\ii}^{2p} + [\bm{\tilde{\epsilon}_k}]_{\jj}^{2p}+[\bm{\tilde{\epsilon}_k}]_{\kk}^{2p}] \\
    =& 3^{p-1}(2p-1)!! [\tilde{\sigma}_{11}^{2p}+\tilde{\sigma}_{22}^{2p}+\tilde{\sigma}_{33}^{2p}]
    \leq  \frac{1}{2}(2p)! [\sqrt{\Tr(\bm{\widetilde{\Sigma}_c}) }]^{2p},
     \nonumber
    \end{aligned}
\end{equation}
where we use $2\times3^{p-1}\leq (2p)!!$ in the last inequality. Therefore, we obtain 
$$||\mathbbm{E}\bm{S_k^{2p}}||_{\op} \leq \frac{1}{2}(2p)![\sqrt{\Tr(\bm{\widetilde{\Sigma}_c}) }]^{2p-2}\frac{\Tr(\bm{\widetilde{\Sigma}_c})}{\min\{d_1,d_2\}},$$
showing that $\bm{S_k}$ satisfies moment constraint (\ref{2.3}) for even numbers under $R= \sqrt{\Tr(\bm{\widetilde{\Sigma}_c}) }$, $\sigma^2 = \frac{R^2}{\min\{d_1,d_2\}}$. Moreover, for $p\geq 1$ we have 
$$\bm{S_k}^{2p+1}=\begin{bmatrix}\bm{0} & |\bm{\tilde{\epsilon}_k}|^{2p}\bm{\tilde{\epsilon}_k}\bm{X_k} \\  |\bm{\tilde{\epsilon}_k}|^{2p}\bm{\tilde{\epsilon}_k^*}\bm{X_k^T} & \bm{0} \end{bmatrix},$$
hence $\mathbbm{E}\bm{S_k}^{2p+1}=\bm{0}$. By using Lemma \ref{lemma4}, for any $t>0$ it holds that 
\begin{equation}
\mathbbm{P}\Big[ ||\frac{1}{n}\sum_{k=1}^n \bm{\tilde{\epsilon}_kX_k} ||_{\op}\geq t\Big]\leq 4(d_1+d_2)\exp \left(-\frac{n\min\{d_1,d_2\}t^2}{2R(R+t\min\{d_1,d_2\})} \right),
    \nonumber
\end{equation}
where $R= \sqrt{\Tr(\bm{\widetilde{\Sigma}_c})}$. We then plug in 
$t = 2R\sqrt{\frac{\log(d_1+d_2)}{n\min\{d_1,d_2\}}},$
assume $\frac{\min\{d_1,d_2\}\log(d_1+d_2)}{n}$ is small, we have 
\begin{equation}
\mathbbm{P}\Big[ ||\frac{1}{n}\sum_{k=1}^n \bm{(W\epsilon_k)X_k} ||_{\op}\geq 2\sqrt{\frac{\Tr(\bm{W\Sigma W})\log(d_1+d_2)}{n\min\{d_1,d_2\}}}\Big]\leq 4(d_1+d_2)^{-3/4}.
    \label{B.25}
\end{equation}
Therefore, with probability higher than $1-4(d_1+d_2)^{-3/4}$, (\ref{3.14}) holds if we choose $\lambda$ by (\ref{3.18}). \hfill $\square$

\vspace{3mm}
\noindent
\textbf{Proof of Theorem \ref{theorem2}.} By (\ref{3.6}), (\ref{3.7}), we have $|| \bm{\widehat{\Delta}}||_{\mathrm{w},\infty} \leq ||\bm{\widehat{\Theta}} ||_{\mathrm{w},\infty}+ || \bm{\widetilde{\Theta}}||_{\mathrm{w},\infty}\leq 2\alpha.$
By Lemma \ref{lemma7} we can rule out probability $1-4(d_1+d_2)^{-3/4}$ and assume (\ref{3.14}) holds, by Lemma \ref{lemma6} we obtain (\ref{3.15}). Then we consider the constraint set $\mathcal{C}(2,\frac{5C_1}{C_1-1},\psi_3),$
where $\psi_3$ is slightly large such that Lemma \ref{lemma5} is valid, i.e., there exists $\kappa \in (0,1)$ such that with probability higher than $1-2(d_1+d_2)^{-3/4}$,
\begin{equation}
\mathcal{F}_{\mathscr{X}}(\bm{\Theta}) \geq \kappa \frac{|| \bm{\Theta}||^2_{\mathrm{w},\mathrm{F}}}{d_1d_2}- T_0,
    \label{B.26}
\end{equation}
holds for all $\bm{\Theta} \in \mathcal{C}(2,5C_1/(C_1-1),\psi_3)$. Therefore, both (\ref{3.15}) and (\ref{B.26}) hold with probability at least $1-6(d_1+d_2)^{-3/4}$. We show (\ref{3.19}) by discussing whether $\bm{\widehat{\Delta}}$ falls onto $\mathcal{C}(2,5C_1/(C_1-1),\psi_3)$.

\vspace{2mm}
\noindent
\textbf{Case 1.} $\bm{\widehat{\Delta}}\in\mathcal{C}(2,5C_1/(C_1-1),\psi_3) $, and $T_0 \leq \frac{\kappa}{2d_1d_2}||\bm{\widehat{\Delta}} ||^2_{\mathrm{w},\mathrm{F}}$. Then by (\ref{B.26}) we have 
\begin{equation}
\mathcal{F}_{\mathscr{X}}(\bm{\widehat{\Delta}}) \geq\frac{\kappa}{2d_1d_2}||\bm{\widehat{\Delta}} ||^2_{\mathrm{w},\mathrm{F}}.
    \label{B.27}
\end{equation}
By combining with (\ref{B.20}), (\ref{B.21}), (\ref{3.14}), we derive an inequality
\begin{equation}
    \begin{aligned}
    \lambda || \bm{\widehat{\Delta}}||_{\nuc} &\geq  \lambda (||\bm{\widetilde{\Theta}} ||_{\nuc}-|| \bm{\widehat{\Theta}}||_{\nuc}) 
    \geq  \mathcal{L}_{\mathrm{w}}(\bm{\widehat{\Theta}}) - \mathcal{L}_{\mathrm{w}}(\bm{\widetilde{\Theta}})  = \frac{1}{2} \mathcal{F}_{\mathscr{X}}(\bm{\widehat{\Delta}} ) - \mathrm{Re}\big<\frac{1}{n}\sum_{k=1}^n (\bm{W\epsilon_k})\bm{X_k}, \bm{\widehat{\Delta}}\big> \\
    &\geq  \frac{\kappa}{4d_1d_2}|| \bm{\widehat{\Delta}}||_{\mathrm{w},\mathrm{F}}^2 - \frac{\lambda}{C_1}  || \bm{\widehat{\Delta}}||_{\nuc} \geq  \frac{\kappa \lambda_{\min}(\bm{W})}{4d_1d_2}|| \bm{\widehat{\Delta}}||_{\mathrm{F}}^2 - \frac{\lambda}{C_1}|| \bm{\widehat{\Delta}}||_{\nuc}.
    \nonumber
    \end{aligned}
\end{equation}
Further plug in (\ref{3.15}), (\ref{3.18}), we obtain 
\begin{equation}
|| \bm{\widehat{\Delta}}||_{\mathrm{F}}^2 \lesssim  \frac{\rho}{(d_1d_2)^{q/2}}\frac{d_1d_2}{\lambda_{\min}(\bm{W})}\rho^{\frac{1}{2-q}}|| \bm{\widehat{\Delta}}||_{\mathrm{F}}^{\frac{2-2q}{2-q}}\sqrt{\frac{\Tr(\bm{W\Sigma_cW})\log(d_1+d_2)}{n\min\{d_1,d_2\}}},
    \nonumber
\end{equation}
by some algebra it reduces to 
\begin{equation}
\frac{||\bm{\widehat{\Delta}} ||_{\mathrm{F}}^2}{d_1d_2} \lesssim \frac{\rho}{(d_1d_2)^{q/2}}\left(\frac{\Tr(\bm{W\Sigma_c\bm{W}})}{\lambda_{\min}(\bm{W})^2} \frac{\max\{d_1,d_2\}\log(d_1+d_2)}{n}\right)^{1-q/2}.
    \label{B.28}
\end{equation}

\vspace{2mm}
\noindent
\textbf{Cases 2.} $\bm{\widehat{\Delta}}\in\mathcal{C}(2,5C_1/(C_1-1),\psi_3) $, and $T_0 \geq \frac{\kappa}{2d_1d_2}||\bm{\widehat{\Delta}} ||^2_{\mathrm{w},\mathrm{F}}$. Then the latter inequality implies $T_0 \geq \frac{\kappa\lambda_{\min}(\bm{W})}{2d_1d_2}|| \bm{\widehat{\Delta}}||^2_{\mathrm{F}}$, we plug in $T_0$ (\ref{3.12}), some algebra yields 
\begin{equation}
\frac{||\bm{\widehat{\Delta}} ||_{\mathrm{F}}^2}{d_1d_2} \lesssim \frac{\rho}{(d_1d_2)^{q/2}}\left(\frac{||\bm{W} ||_{\infty}\alpha^2}{\lambda_{\min}(\bm{W})^3}\frac{\max\{d_1,d_2\}\log(d_1+d_2)}{n}\right)^{1-q/2}.
    \label{B.29}
\end{equation}

\vspace{2mm}
\noindent
\textbf{Cases 3.} $\bm{\widehat{\Delta}}\notin\mathcal{C}(2,5C_1/(C_1-1),\psi_3) $. We have showed $\bm{\widehat{\Delta}}$ satisfies the first two constraints of $\mathcal{C}(2,5C_1/(C_1-1),\psi_3)$, so it violates the third one, i.e., 
$|| \bm{\widehat{\Delta}}||_{\mathrm{w},\mathrm{F}}^2 \leq \alpha^2(d_1d_2)\sqrt{\frac{\psi_3 \log(d_1+d_2)}{n}},$
combine with $||\bm{\widehat{\Delta}}||_{\mathrm{w},\mathrm{F}}^2 \geq \lambda_{\min}(\bm{W})||\bm{\widehat{\Delta}}||_{\mathrm{F}}^2$, we obtain 
\begin{equation}
\frac{|| \bm{\widehat{\Delta}}||_{\mathrm{F}}^2}{d_1d_2} \leq \frac{\alpha^2}{\lambda_{\min}(\bm{W})} \sqrt{\frac{\psi_3 \log(d_1+d_2)}{n}},
    \label{B.30}
\end{equation}
which is also dominated by the right hand side of (\ref{3.19}) under the scaling $n<d_1d_2$ and $\rho \gtrsim (d_1d_2)^{q/2}$. So the result follows. \hfill $\square$

\subsection{Noise Correction}
\textbf{Proof of Lemma \ref{lemma8}.} Let the eigenvalue decomposition of $\bm{\Sigma_c}$ be $\bm{O\Lambda O^T}$, where $\bm{O}$ is orthogonal matrix, $\bm{\Lambda} = \mathrm{diag}(\lambda_1, \lambda_2,\lambda_3)$. We then define $\bm{\widetilde{W}} =\bm{O^TWO}$, which is positive definite with trace equaling 3, and we use $\tilde{w}_{ij}$ to denote the (i,j) entry of $\bm{\widetilde{W}}$, so the constraint is $\tilde{w}_{11}+\tilde{w}_{22}+\tilde{w}_{33}=3$. Then we plug in the objective, it yields that 
\begin{equation}
    \begin{aligned}
    &\Tr(\bm{W\Sigma_c W})  
    = \Tr(\bm{\widetilde{W}\widetilde{W}^T\Lambda}) 
    = \sum_{i=1}^3 \lambda_i [ \sum_{j=1}^3 \tilde{w}_{ij}^2] 
    \geq  \sum_{i=1}^3 \lambda_i \tilde{w}_{ii}^2 
    \geq  \frac{9}{\lambda_1^{-1}+ \lambda_2^{-1}+ \lambda_3^{-1}}\ ,
    \nonumber
    \end{aligned}
\end{equation}
where the first inequality is obvious, and the equal sign holds if and only if $\bm{\widetilde{W}}$ is diagonal, and the second inequality can be verified by KKT condition, and the equal sign is true if and only if $ \tilde{w}_{ii}= \frac{3\lambda_i^{-1}}{\sum_{i=1}^3 \lambda_i^{-1}},\ \forall 1\leq i\leq 3. $
Therefore, the unique $\bm{W}$ that minimizes $\Tr(\bm{W\Sigma_c W})$ is $\frac{3}{\Tr(\bm{\Sigma_c^{-1}})}\bm{O}\mathrm{diag}(\lambda_1^{-1},\lambda_2^{-1},\lambda_3^{-1}) \bm{O^T}$.
 \hfill $\square$
 
\end{document}